\documentclass[11pt, a4paper]{preprint_style/style}
\usepackage[sort&compress]{natbib}
\usepackage{changepage}

\usepackage{microtype}
\usepackage{graphicx}
\usepackage{subcaption}
\usepackage{booktabs} % for professional tables
\usepackage{hyperref}
\usepackage{array} %新加的
\usepackage{tabularx}
\usepackage{pifont}
\usepackage{makecell}
\usepackage{multirow} 
\usepackage{xcolor}          % 颜色支持
\usepackage{float}
\usepackage{enumitem}
\usepackage{amsmath}
\usepackage{amssymb}
\usepackage{mathtools}
\usepackage{amsthm}
\usepackage{placeins}
\usepackage{pifont}   % for \ding
\usepackage{enumitem}
\usepackage[capitalize,noabbrev]{cleveref}
\usepackage{amsmath,amsfonts,bm}

\def\eqref#1{equation~\ref{#1}}
\def\1{\bm{1}}

\DeclareMathAlphabet{\mathsfit}{\encodingdefault}{\sfdefault}{m}{sl}
\SetMathAlphabet{\mathsfit}{bold}{\encodingdefault}{\sfdefault}{bx}{n}

\newcommand{\LongTask}{Long Horizon Task}
\newcommand{\APPGroup}{Functional App Group}
\newcommand{\methodname}{Anchored State Memory}
\newcommand{\methodshortname}{ASM}
\newcommand{\StateAnchors}{State Anchors}
\newcommand{\BenchName}{AndroTMem}
\newcommand{\cmark}{\textcolor{green!70!black}{\ding{51}}} % 绿色对号
\newcommand{\xmark}{\textcolor{red!70!black}{\ding{55}}}   % 红色叉号

\newcommand{\cmarkbf}{\textbf{\textcolor{green!70!black}{\ding{51}}}}
\newcommand{\act}[1]{\textsc{#1}}
\newcommand{\Dependency}{cross-step causal dependencies}
\theoremstyle{plain}

\theoremstyle{definition}

\theoremstyle{remark}

\usepackage[textsize=tiny]{todonotes}

\usepackage[utf8]{inputenc} %
\usepackage[T1]{fontenc}    %
\usepackage{hyperref}       %
\usepackage{url}            %
\usepackage{tabularx}
\usepackage{makecell}
\usepackage{array,ragged2e,booktabs}
\newcolumntype{P}[1]{>{\RaggedRight\arraybackslash}p{#1}}
\usepackage{longtable}
\usepackage{amsfonts}       %
\usepackage{amsthm}         %
\usepackage{nicefrac}       %
\usepackage{microtype}      %
\usepackage{xcolor}         %

\usepackage{algorithm}
\usepackage{algorithmicx}
\usepackage{algpseudocode}

\definecolor{darkblue}{rgb}{0, 0, 0.5}
\hypersetup{colorlinks=true, citecolor=darkblue, linkcolor=darkblue, urlcolor=darkblue}

\usepackage{xurl}

\usepackage{soul}
\usepackage{amsmath}
\usepackage{subcaption}
\usepackage{graphicx}
\usepackage{changes}
\usepackage{amsmath,mathtools}
\usepackage{changes}
\usepackage{pifont}
\usepackage{tocloft}

\usepackage{lipsum}
\usepackage{colortbl}
\usepackage{wrapfig}
\usepackage{xspace}
\usepackage{multirow}
\usepackage{enumitem}

\usepackage{pifont}
\usepackage{listings}
\usepackage{fontawesome5} 

\usepackage{wrapfig}
\usepackage{caption}

\usepackage{makecell}

\usepackage{tabularx}
\usepackage{afterpage}

\usepackage[edges]{forest}
\usepackage[normalem]{ulem}
\usepackage{caption}
\usepackage{CJKutf8}
\usepackage{bbding}
\usepackage[most,skins,theorems]{tcolorbox}
\usepackage{booktabs}
\usepackage{multirow}
\usepackage{hyperref}
\usepackage{array}

\usepackage[tikz]{bclogo}
\usepackage[framemethod=tikz]{mdframed}
\definecolor{bgblue}{RGB}{245,243,253}
\definecolor{ttblue}{RGB}{91,194,224}

\mdfdefinestyle{mystyle}{%
  rightline=true,
  innerleftmargin=10,
  innerrightmargin=10,
  outerlinewidth=3pt,
  topline=false,
  rightline=true,
  bottomline=false,
  skipabove=\topsep,
  skipbelow=\topsep
}

\newtcolorbox{myboxi}[1][]{
  breakable,
  title=#1,
  colback=red!5,
  colbacktitle=red!5,
  coltitle=black,
  fonttitle=\bfseries,
  bottomrule=0pt,
  toprule=0pt,
  leftrule=2pt,
  rightrule=2pt,
  titlerule=0pt,
  arc=0pt,
  outer arc=0pt,
  colframe=red,
}

\newtcolorbox{myboxnote}[1][]{
  breakable,
  title=#1,
  colback=orange!0,
  colbacktitle=orange!0,
  coltitle=black,
  fonttitle=\bfseries,
  bottomrule=0pt,
  toprule=0pt,
  leftrule=2pt,
  rightrule=2pt,
  titlerule=0pt,
  arc=0pt,
  outer arc=0pt,
  colframe=orange,
}

\usepackage{epigraph}

\newtcolorbox{myboxii}[1][]{
  breakable,
  freelance,
  title=#1,
  colback=white,
  colbacktitle=white,
  coltitle=black,
  fonttitle=\bfseries,
  bottomrule=0pt,
  boxrule=0pt,
  colframe=white,
  overlay unbroken and first={
  \draw[red!75!black,line width=3pt]
    ([xshift=5pt]frame.north west) -- 
    (frame.north west) -- 
    (frame.south west);
  \draw[red!75!black,line width=3pt]
    ([xshift=-5pt]frame.north east) -- 
    (frame.north east) -- 
    (frame.south east);
  },
  overlay unbroken app={
  \draw[red!75!black,line width=3pt,line cap=rect]
    (frame.south west) -- 
    ([xshift=5pt]frame.south west);
  \draw[red!75!black,line width=3pt,line cap=rect]
    (frame.south east) -- 
    ([xshift=-5pt]frame.south east);
  },
  overlay middle and last={
  \draw[red!75!black,line width=3pt]
    (frame.north west) -- 
    (frame.south west);
  \draw[red!75!black,line width=3pt]
    (frame.north east) -- 
    (frame.south east);
  },
  overlay last app={
  \draw[red!75!black,line width=3pt,line cap=rect]
    (frame.south west) --
    ([xshift=5pt]frame.south west);
  \draw[red!75!black,line width=3pt,line cap=rect]
    (frame.south east) --
    ([xshift=-5pt]frame.south east);
  },
}

\definecolor{myblue}{rgb}{0.9, 0.1, 0.94}
\definecolor{mygreen}{rgb}{0.64, 0.56, 0.88}
\definecolor{myyellow}{rgb}{0.68, 0.6, 0.1}
\definecolor{fancygreen}{rgb}{0.33, 0.68, 0.20}
\definecolor{salmon}{rgb}{0.94, 0.52, 0.49}
\definecolor{tablegreen}{rgb}{0.82, 0.94, 0.75}
\definecolor{tableblue}{rgb}{0.81, 0.90, 0.94}
\definecolor{tablered}{rgb}{0.97, 0.85, 0.85}
\definecolor{tableorange}{rgb}{0.96, 0.85, 0.81}

\definecolor{myorange}{rgb}{1.0, 0.49, 0.0}	
\definecolor{tlgreen}{rgb}{0.33, 0.68, 0.20}

\newenvironment{itemize*}%
 {\leftmargini=10pt\begin{itemize}%
  \setlength{\itemsep}{0pt}%
  \setlength{\parskip}{0pt}%
  }%
 {\end{itemize}}
\newenvironment{enumerate*}%
 {\begin{enumerate}%
  \setlength{\itemsep}{0pt}%
  \setlength{\parskip}{0pt}}%
 {\end{enumerate}}

\tikzset{%
    every node/.style={font=\tiny},
    parent/.style =          {align=center,text width=2cm,rounded corners=3pt, line width=0.3mm, fill=gray!10,draw=gray!80},
    child/.style =           {align=center,text width=2.0cm,rounded corners=3pt, fill=blue!10,draw=blue!80,line width=0.3mm},
    grandchild/.style =      {align=center,text width=2cm,rounded corners=3pt},
    greatgrandchild/.style = {align=center,text width=1.5cm,rounded corners=3pt},
    greatgrandchild2/.style = {align=center,text width=1.5cm,rounded corners=3pt},    
    referenceblock/.style =  {align=center,text width=1.5cm,rounded corners=2pt},
    pretrain/.style =           {align=center,text width=2.0cm,rounded corners=3pt, fill=blue!10,draw=blue!80,line width=0.3mm},   
    pretrain_work/.style =           {align=center, text width=8.5cm,rounded corners=3pt, fill=blue!10,draw=blue!0,line width=0.3mm},  
    template/.style =           {align=center,text width=2.0cm,rounded corners=3pt, fill=red!10,draw=red!80,line width=0.3mm},   
    template_work/.style =           {align=center,text width=8.5cm,rounded corners=3pt, fill=red!10,draw=red!0,line width=0.3mm},    
    answer/.style =           {align=center,text width=2.0cm,rounded corners=3pt, fill= cyan!10,draw= cyan!80,line width=0.3mm},   
    answer_work/.style =           {align=center,text width=8.5cm,rounded corners=3pt, fill= cyan!10,draw= cyan!0,line width=0.3mm},      
    multiple/.style =           {align=center,text width=2.0cm,rounded corners=3pt, fill= orange!10,draw= orange!80,line width=0.3mm},   
    multiple_work/.style =           {align=center,text width=8.5cm,rounded corners=3pt, fill= orange!10,draw= orange!0,line width=0.3mm},        
    tuning/.style =           {align=center,text width=2.0cm,rounded corners=3pt, fill= magenta!10,draw= magenta!80,line width=0.3mm},   
    tuning_work/.style =           {align=center,text width=8.5cm,rounded corners=3pt, fill= magenta!10,draw= magenta!0,line width=0.3mm},          
}

\usepackage{tikz}

\newcommand{\lstbg}[3][0pt]{{\fboxsep#1\colorbox{#2}{\strut #3}}}

\lstdefinelanguage{diff}{
  basicstyle=\ttfamily\small,
  morecomment=[f][\lstbg{red!20}]-,
  morecomment=[f][\lstbg{green!20}]+,
}
\lstdefinelanguage{diffpython}{
  language=diff,
  morekeywords={def, if, else, for, while, return, import, from, as, class, with, try, except, finally, raise, lambda, and, or, not, in, is, None, True, False},
  morecomment=[l]{\#},
  morestring=[b]",
  morestring=[b]',
}

\definecolor{darkgreen}{RGB}{50,100,0}
\definecolor{darkred}{RGB}{200, 0, 0}
\definecolor{lightblue}{RGB}{220,235,250}

\tcbset{
  takeawaysbox/.style={
    title=Takeaways,
    colback=lightblue!80,
    colframe=black,
    fonttitle=\bfseries\small,
    coltitle=white,
    colbacktitle=black,
    enhanced,
    attach boxed title to top left={xshift=2.5mm,yshift=-2.5mm},
    boxed title style={rounded corners, size=small, colframe=black, colback=black},
    width=\linewidth,
    arc=3.5mm
  }
}

\definecolor{darkgreen}{RGB}{50,100,0}
\definecolor{darkred}{RGB}{200, 0, 0}

\NewDocumentCommand{\kaiyan}
{ mO{} }{\textcolor{purple}{\textsuperscript{\textit{kaiyan}}\textsf{\textbf{\small[#1]}}}}
\NewDocumentCommand{\yuxin}
{ mO{} }{\textcolor{cyan}{\textsuperscript{\textit{yuxin}}\textsf{\textbf{\small[#1]}}}}
\NewDocumentCommand{\bx}
{ mO{} }{\textcolor{green}{\textsuperscript{\textit{bx}}\textsf{\textbf{\small[#1]}}}}
\NewDocumentCommand{\at}
{ mO{} }{\textcolor{red}{\textsuperscript{\textit{AT}}\textsf{\textbf{\small[#1]}}}}
\NewDocumentCommand{\re}
{ mO{} }{\textcolor{blue}{\textsuperscript{\textit{RE}}\textsf{\textbf{\small[#1]}}}}
\NewDocumentCommand{\ybsun}
{ mO{} }{\textcolor{magenta}{\textsuperscript{\textit{youbang}}\textsf{\textbf{\small[#1]}}}}
\NewDocumentCommand{\runze}
{ mO{} }{\textcolor{orange}{\textsuperscript{\textit{runze}}\textsf{\textbf{\small[#1]}}}}

\definecolor{darkgreen}{RGB}{0,100,0} 
\NewDocumentCommand{\add}
{ mO{} }{\textcolor{darkgreen}{\textsuperscript{\textit{Maybe Consider Discuss}}\textsf{\textbf{[#1]}}}}

\setlist[itemize]{leftmargin=20pt}

\usepackage[edges]{forest}
\definecolor{hidden-blue}{RGB}{194,232,247}
\definecolor{hidden-black}{RGB}{20,68,106}

\usepackage{tabularx, booktabs, xcolor, colortbl}
\usepackage{pifont}   %
\usepackage{xstring}  %

\usepackage{multirow}
\usepackage{hyperref}
\usepackage{fontawesome5}

\usepackage[export]{adjustbox}

\definecolor{yes}{HTML}{C6EFCE}      %
\definecolor{no}{HTML}{FFC7CE}       %
\definecolor{partial}{HTML}{FFEB9C}  %
\definecolor{external}{HTML}{D9E1F2} %
\definecolor{hdr}{HTML}{F2F2F2}

\usepackage{amssymb}

\newcommand{\cellstatus}[1]{%
  \begingroup
  \StrTrim{#1}[\statusval]%
  \IfStrEq{\statusval}{Yes}{\cellcolor{yes}\cmark}{}%
  \IfStrEq{\statusval}{No}{\cellcolor{no}\xmark}{}%
  \IfBeginWith{\statusval}{Yes (}{\cellcolor{yes}\cmark~\textit{\statusval\unskip}}{}%
  \IfStrEq{\statusval}{Partial}{\cellcolor{partial}\textbf{Partial}}{}%
  \IfStrEq{\statusval}{External}{\cellcolor{external}\textbf{External}}{}%
  \endgroup
}

\title{AndroTMem: From Interaction Trajectories to Anchored Memory in Long-Horizon GUI Agents}

\author{%
  { 
    Yibo Shi$^{1,*}$,~
    Jungang Li$^{2,3,*,\dagger}$,~
    Linghao Zhang$^{4,*}$,~
    Zihao Dongfang$^{2,3,*}$,~
    Biao~Wu$^{5}$,~
    Sicheng Tao$^{2}$,
    Yibo~Yan$^{2,3}$,~
    Chenxi Qin$^{6}$,~
    Weiting Liu$^{7}$,~
    Zhixin Lin$^{8}$,~
    Hanqian Li$^{2}$,~
    Yu Huang$^{2}$,~
    Song~Dai$^{2,3}$,
    Yonghua~Hei$^{2,3}$,
    Yue Ding$^{9}$,~
    Xiang Li$^{2,3}$,~
    Shikang Wang$^{4}$,~
    Chengdong Xu$^{10}$,~
    Jingqi Liu$^{1}$,
    Xueying~Ma$^{1}$,~
    Zhiwen~Zheng$^{1}$,
    Xiaofei~Zhang$^{1}$,~
    Bincheng Wang$^{11}$,~
    Nichen Yang$^{1}$,~
    Jie Wu$^{10}$,~
    Lihua~Tian$^{1,\ddagger}$,~
    Chen~Li$^{1,\ddagger}$,~
    Xuming Hu$^{2,3,\ddagger}$\par
  }

  \vspace{1mm}

  $^{1}$ XJTU \quad
  $^{2}$ HKUST(GZ) \quad
  $^{3}$ HKUST \quad
  $^{4}$ CityU \quad
  $^{5}$ UTS \quad
  $^{6}$ TJU \quad
  $^{7}$ FDU \quad
  $^{8}$ SDU \quad
  $^{9}$ CASIA \quad
  $^{10}$ SYSU \quad
  $^{11}$~NWPU
  \vspace{1mm} \\

  $^*$ \textbf{Core Contributors.} \quad
  $^\dagger$ \textbf{Project Lead.} \quad
  $^\ddagger$ \textbf{Corresponding Authors.}
  \vspace{1mm} \\
  \textbf{Contact:} \faEnvelope[regular]~\texttt{yiboshi86@gmail.com}, \texttt{ljungang.02@gmail.com}, \texttt{lhtian@xjtu.edu.cn}, \texttt{xuminghu@hkust-gz.edu.cn}
}

\begin{abstract}
Long-horizon GUI agents are a key step toward real-world deployment, yet effective interaction memory under prevailing paradigms remains under-explored. Replaying full interaction sequences is redundant and amplifies noise, while summaries often erase dependency-critical information and traceability. We present \textbf{AndroTMem}, a diagnostic framework for anchored memory in long-horizon Android GUI agents. 
Its core benchmark, \textbf{\BenchName{}-Bench}, comprises 1,069 tasks with 34,473 interaction steps (avg. 32.1 per task, max. 65).
We evaluate agents with \textbf{TCR} (Task Complete Rate), focusing on tasks whose completion requires carrying forward critical intermediate state; AndroTMem-Bench is designed to enforce strong step-to-step causal dependencies, making sparse yet essential intermediate states decisive for downstream actions and centering interaction memory in evaluation.
Across open- and closed-source GUI agents, we observe a consistent pattern: as interaction sequences grow longer, performance drops are driven mainly by within-task memory failures, not isolated perception errors or local action mistakes.
Guided by this diagnosis, we propose \textbf{Anchored State Memory (ASM)}, which represents interaction sequences as a compact set of causally linked intermediate-state anchors to enable subgoal-targeted retrieval and attribution-aware decision making. Across multiple settings and 12 evaluated GUI agents, ASM consistently outperforms full-sequence replay and summary-based baselines, improving TCR by 5\%--30.16\% and AMS by 4.93\%--24.66\%, indicating that anchored, structured memory effectively mitigates the interaction-memory bottleneck in long-horizon GUI tasks. The code, benchmark, and related resources are publicly available at \url{https://github.com/CVC2233/AndroTMem}.

\keywords{Smartphone GUI Agent, Long-Horizon Task, Agent Memory Evaluation, Anchored State Memory }
\end{abstract}

\begin{document}

\maketitle

% \begin{figure}[h]
% \centering
% \includegraphics[width=\textwidth]{figures/teaser2.pdf}
% \caption{
% Overview of the survey.
% }
% \label{fig:teaser}
% \end{figure}

% \newpage
% \begingroup
% \setlength{\baselineskip}{1.25\baselineskip}
% \tableofcontents
% \endgroup

% \newpage

\section{Introduction}
Multimodal large language models (MLLMs) have rapidly progressed from static image understanding~\cite{wang2024gui,dang-etal-2025-exploring,ding2026omnisift,zheng2026visual} toward richer forms of perception and interaction, including stronger visual reasoning~\cite{dong2025insight}, video understanding~\cite{tang2025video,liu2025javisgpt,tao2025moss,hu2025videomark,pan2026egointent}, and increasingly responsive real-time multimodal interfaces~\cite{comanici2025gemini,hurst2024gpt,bai2025qwen2,xun2025rtv}. This evolution is reshaping human--computer collaboration: systems are moving beyond tools that execute isolated commands toward agents that can perceive, reason, and act continuously in real-world environments~\cite{zhang2024large,li2025survey}. Along this trajectory, GUI agents have emerged as one of the most practical and impactful instantiations, aiming to complete everyday and professional tasks from natural-language instructions across heterogeneous device settings such as mobile, PC, and web~\cite{rawles2024androidworld,rawles2023androidinthewild,liu2025pc,zheng2024gpt,xie2024osworld,huang2025hyperg}.

% fig:1_teaser
\begin{figure}[t]
    \centering
    \includegraphics[width=\linewidth]{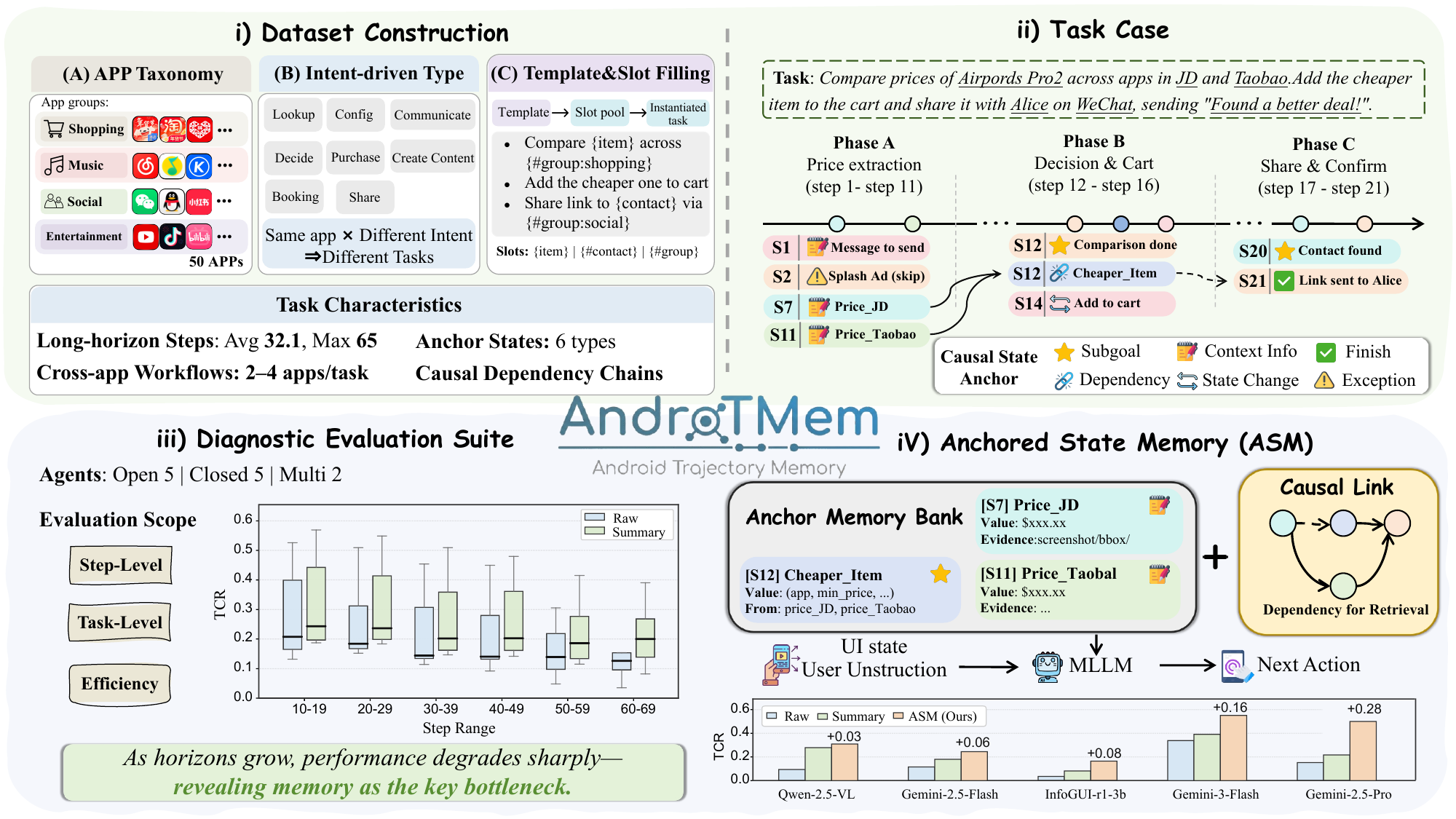}
    \caption{Overview of \textbf{AndroTMem}. AndroTMem comprises: 
\textbf{(i)} \textbf{AndroTMem-Bench}, a long-horizon Android GUI benchmark constructed with intent-driven, cross-app, causally dependent workflows; 
\textbf{(ii)} representative task cases where sparse intermediate states determine downstream decisions; 
\textbf{(iii)} a diagnostic evaluation suite showing that performance drops with horizon length are primarily caused by memory failures; and 
\textbf{(iv)} \textbf{Anchored State Memory (ASM)}, which stores causally linked intermediate-state anchors for targeted retrieval and improves long-horizon GUI-agent performance.
}
    \label{fig:1_teaser}
    % \vspace{-2em}
\end{figure}

Despite steady progress in UI grounding and single-step action reasoning~\cite{qin2025ui,cheng2024seeclick,gou2024navigating,wu2024atlas,lin2025showui,li2025screenspot}, GUI agents remain brittle when interaction trajectories extend to dozens of steps. Real-world mobile tasks are rarely just long chains of loosely connected actions; instead, they consist of indispensable intermediate steps whose outcomes must be preserved and reused later. Users often need to extract values across pages, verify prerequisite conditions, handle exception branches, and bring these earlier results back when they become relevant several steps later. As a result, the central challenge in long-horizon execution is no longer only perceiving the current interface or selecting the next action, but maintaining, retrieving, and operationalizing critical intermediate state over time---that is, \emph{anchored memory}.

Recent studies have begun to explore how memory mechanisms can improve planning and reasoning in agentic systems~\cite{park2023generativeagents,packer2023memgpt,shinn2023reflexion}. However, mainstream GUI-agent benchmarks still focus primarily on short-horizon routines or weakly coupled multi-step tasks, and thus lack dedicated evaluation of agents' memory ability in long-horizon tasks~\cite{rawles2023androidinthewild,chai2025amex,chen2024spa,wang2024mobileagentbench,xu2025mobile}. In such settings, later decisions may succeed even without faithfully reusing earlier information, which obscures a more fundamental issue in real workflows: many intermediate results that determine downstream success---such as extracted values, prerequisite completions, or exception-handling outcomes---may exert their causal effect only several steps later. Existing evaluations therefore expose a critical blind spot: they can measure whether an agent completes a workflow, but not whether it truly preserves and correctly reuses task-critical state across time~\cite{wang2025mmbench,liu2025verigui}.

In practice, this blind spot manifests as three tightly coupled challenges. First, existing datasets are not designed to evaluate long-horizon memory explicitly: their length often comes from chaining loosely related steps rather than enforcing strong step-to-step causal dependencies. Second, long-horizon failure is poorly diagnosed: end-to-end success conflates perception errors, local action mistakes, and memory breakdowns, offering little visibility into where degradation begins and what drives it as trajectories lengthen. Third, history modeling remains at an impasse: replaying full interaction sequences is redundant and amplifies noise, while coarse summaries often erase exactly the fine-grained states and dependencies required downstream. Together, these gaps obscure genuine progress in long-horizon GUI agents and make memory bottlenecks difficult to measure and improve reliably~\cite{wang2024gui,zhang2024large}.

To address this, we present \textbf{AndroTMem}, a diagnostic framework for interaction memory in long-horizon Android GUI agents. At its core is \textbf{AndroTMem-Bench}, comprising 1,069 realistic tasks and 34,473 interaction steps across 50 applications (avg. 32.1 steps per task, max. 65). The benchmark is explicitly constructed to enforce strong step-to-step causal dependencies, making sparse but essential intermediate states decisive for downstream decisions. We evaluate agents with \textbf{TCR} (Task Complete Rate), focusing on key tasks whose completion hinges on carrying forward critical intermediate state, thereby placing long-horizon memory ability at the center of evaluation. Systematic studies across diverse open- and closed-source GUI agents reveal a consistent pattern: as interaction sequences lengthen, performance drops are driven primarily by \emph{within-task memory failures}, rather than isolated perception errors or local action mistakes.

Our main contributions are summarized as follows:
\ding{182} We introduce \textbf{AndroTMem}, a diagnostic framework for memory in long-horizon GUI agents, together with \textbf{AndroTMem-Bench}, a benchmark that evaluates memory via \textbf{TCR} on dependency-critical long-horizon tasks.
\ding{183} Using this benchmark, we show that long-horizon degradation is dominated by \emph{within-task memory failures}, not isolated perception errors or local action mistakes.
\ding{184} We propose \textbf{Anchored State Memory (ASM)}, which organizes history into causally linked intermediate-state anchors for targeted retrieval and attribution.

% \begin{itemize}
%     \item[\ding{182}] We introduce \textbf{AndroTMem}, a diagnostic framework for memory in long-horizon GUI agents, together with \textbf{AndroTMem-Bench}, a benchmark that evaluates memory via \textbf{TCR} on dependency-critical long-horizon tasks.
%     \item[\ding{183}] Using this benchmark, we show that long-horizon degradation is dominated by \emph{within-task memory failures}, not isolated perception errors or local action mistakes.
%     \item[\ding{184}] We propose \textbf{Anchored State Memory (ASM)}, which organizes history into causally linked intermediate-state anchors for targeted retrieval and attribution.
% \end{itemize}
Across multiple settings and 12 GUI agents, ASM consistently outperforms full-sequence replay and summary-based baselines, improving TCR by 5\%--30.16\% and AMS by 4.93\%--24.66\%, underscoring anchored memory as a core capability for reliable and scalable long-horizon GUI agents.
\section{Related Work}
\label{sec:2_related_work}

\paragraph{Long-Horizon Task Execution and Memory in GUI Agents.}
GUI agents aim to complete user instructions by iteratively perceiving the current interface and issuing low-level actions such as taps, swipes, and text input. Recent MLLMs have substantially improved UI understanding and instruction following, enabling stronger page perception, UI grounding, and step-level action prediction~\cite{zhang2024large,wang2024gui,chen2025knowmt}. Prior work has therefore progressed from improving perception and element grounding~\cite{hong2024cogagent,gou2024navigating,cheng2024seeclick,lin2025mind} to studying multi-step GUI interaction through end-to-end agents~\cite{lin2025showui,wu2024atlas}, modular systems~\cite{liu2025learnact,wang2025mobile}, and diverse training strategies including supervised learning, reinforcement learning, and self-improvement~\cite{li2025survey,lu2025ui}. 

Despite these advances, reliably executing long-horizon GUI tasks remains challenging. In realistic workflows, interaction trajectories often span dozens of steps across multiple applications, where later decisions depend on intermediate results obtained several steps earlier, such as extracted values, completed subgoals, or environment changes. This makes effective history utilization and intermediate-state management essential for successful task execution~\cite{gu2025ui,liu2025pc,xi2025agentgym}. However, existing GUI-agent frameworks incorporate history through raw interaction traces~\cite{wu2024atlas,lin2025showui,lu2025guiodyssey}, compressed summaries~\cite{liu2025learnact,lu2025ui,xu2025androidlab}, or other generic context aggregation strategies. While these mechanisms help incorporate past information, they are not explicitly designed to preserve and retrieve sparse but causally critical intermediate states in long-horizon cross-app workflows. This limitation motivates memory mechanisms that explicitly model and retrieve decision-critical intermediate states during long-horizon GUI interaction.

\paragraph{Benchmarks and Datasets for Mobile GUI Agents.}
% GUI agent research mainly focuses on perception and action grounding.
% long-horizon interaction is still brittle
% SeeClick, CogAgent, ShowUI, UI-TARS, OS-Atlas
Benchmarks for GUI agents have evolved from single-screen or few-step operations~\cite{chen2024gui,xie2024osworld} to longer multi-step tasks emphasizing end-to-end completion~\cite{liu2025verigui,wang2025mmbench}, with evaluation environments becoming increasingly realistic and covering more diverse web and mobile applications~\cite{kong2025mobileworld,xu2025androidlab}. In parallel, agent observations have shifted from structured UI metadata toward multimodal and pure-vision settings, improving generality and reducing dependence on platform-specific annotations~\cite{lu2025guiodyssey,li2025screenspot}. However, existing benchmarks still provide limited support for studying long-horizon interaction memory. Under such benchmark settings, some tasks remain relatively short or weakly coupled, such that later decisions may rely primarily on local perception and action prediction rather than on faithfully preserving earlier intermediate outcomes. Consequently, they offer limited visibility into whether agents can maintain and correctly reuse task-critical state across time. This motivates benchmarks that explicitly enforce strong step-to-step causal dependencies and make intermediate state management a first-class target of evaluation.

In contrast to prior work, \BenchName{} focuses on explicitly diagnosing interaction memory in long-horizon GUI agents. Our benchmark enforces strong cross-step causal dependencies, enabling systematic evaluation of how agents preserve and reuse intermediate states. Furthermore, we propose ASM, a structured history representation that organizes interaction trajectories around causally linked state anchors.

\section{Dataset Construction and Statistics}
\label{3_preliminaries}
% tab:1_dataset_comparison
\begin{table*}[ht]
\vspace{-6pt}
\centering
\caption{\textbf{Comparison of AndroTMem-Bench with other mobile GUI agent benchmarks.} \#Traj: number of trajectories (tasks); \#Steps: average interaction steps per trajectory; \#App: number of applications. GT: availability of ground-truth trajectories; Low-Level Instr.: low-level instructions. History Modeling: supported within-task history representations.
}
\label{tab:1_dataset_comparison}
\small
\setlength{\tabcolsep}{6pt}

% \begin{adjustbox}{width=0.90\textwidth}
\resizebox{\linewidth}{!}{
\begin{tabular}{
>{\centering\arraybackslash}m{5cm}
>{\centering\arraybackslash}m{1.2cm}
>{\centering\arraybackslash}m{1.2cm}
>{\centering\arraybackslash}m{1.2cm}
>{\centering\arraybackslash}m{1.2cm} % #GT
>{\centering\arraybackslash}m{2.2cm}
>{\centering\arraybackslash}m{1.6cm}
>{\centering\arraybackslash}m{1.6cm}
}
\toprule
\multirow{2}{*}{\normalsize Dataset} &
\multirow{2}{*}{\normalsize \#Traj} &
\multirow{2}{*}{\normalsize \#Steps} &
\multirow{2}{*}{\normalsize \#App} &
\multirow{2}{*}{\normalsize GT} &
\multirow{2}{*}{\normalsize Low-Level Instr.} &
\multicolumn{2}{c}{\normalsize History Modeling} \\
\cmidrule(l{0.9em}r{0.1em}){7-8}
\small
 &  &  &  &  &  & Summary & ASM \\
\midrule
\normalsize
AMEX~\cite{chai2025amex} & 2,946 & 12.8 & 110 & \cmark & \xmark & \xmark & \xmark \\
AITW~\cite{rawles2023androidinthewild} & 30,378 & 6.5 & 357 & \cmark & \xmark & \xmark & \xmark \\
AITZ~\cite{zhang2024android} & 2,504 & 7.5 & 70 & \cmark & \cmark & \cmark & \xmark \\
Android Control~\cite{li2024effects} & 15,283 & 5.5 & 833 & \cmark & \cmark & \cmark & \xmark \\
Android World~\cite{rawles2024androidworld} & 116 & -- & 20 & \xmark & \xmark & \xmark & \xmark \\
Mobile-R1~\cite{gu2025mobile} & 4,635 & 5.3 & 28 & \cmark & \cmark & \xmark & \xmark \\
Mobile-Bench-v2~\cite{xu2025mobile} & 12,856 & 7.28 & 49 & \xmark & \xmark & \xmark & \xmark \\
GUIOdyssey ~\cite{lu2025guiodyssey} & 8,334 & 15.3 & 212 & \xmark & \cmark & \cmark & \xmark \\
SPA-Bench~\cite{chen2024spa} & 201 & 8.2 & 21 & \xmark & \xmark & \xmark & \xmark \\
LearnGUI~\cite{liu2025learnact} & 2,353 & 13.2 & 73 & \cmark & \cmark & \cmark & \xmark \\
% SPA-Bench~\cite{chen2024spa} & 201 & 8.2 & 21 & \xmark & \xmark & \xmark & \xmark \\
% LearnGUI~\cite{liu2025learnact} & 2,353 & 13.2 & 73 & \cmark & \cmark & \cmark & \xmark \\
LongHorizonUI~\cite{kang2026longhorizonui} & 354 & 22.1 & 28 & \cmark & \cmark & \cmark & \xmark \\
\midrule
\textbf{AndroTMem-Bench} & \textbf{1,069} & \textbf{32.1} & \textbf{50}
& \cmarkbf & \cmarkbf & \cmarkbf & \cmarkbf \\
\bottomrule
\end{tabular}}
\vspace{-6pt}

% \end{adjustbox}
\end{table*}

\subsection{Long-Horizon Task Formulation}
\label{sec:3_long_task_formulation}
% 对于\LongTask的结构化定义，下一段数据收集管线需要用到
% 将其定义为----强因果依赖的多步决策问题

% We formulate a long-horizon GUI task as a sequential decision-making problem \textbf{with strong cross-step causal dependencies}.
% Formally, a task is represented as an interaction trajectory
% \begin{equation}
% \tau = \{(s_0, a_0), (s_1, a_1), \ldots, (s_T, a_T)\},
% \end{equation}
% where $\tau$ denotes a trajectory consisting of a sequence of state--action pairs $(s_t,a_t)$, with $s_t$ representing the GUI state at time step $t$ and $a_t$ the action executed in that state. At each time step $t$, the GUI state $s_t$ is a structured observation capturing the current interface configuration.
% 至关重要的是，与短时程或松散耦合的 GUI 任务不同，\BenchName{} 中的任务是\emph{经过精心构建和标注的，以展现显著的跨步骤因果依赖关系}。
% 后续动作通常取决于早期步骤产生的中间结果（我们称其为中间状态-加一下术语的格式），这些中间状态既无法从初始指令中直接观察到，也无法仅从当前的 GUI 状态中恢复。

% 因此，在 \BenchName{} 中执行任务需要智能体在较长的交互周期内正确建立和维护潜在的任务相关状态，使得有效利用交互历史和中间状态成为任务成功的先决条件，而不是可选的增强功能。（这段话感觉有点冗余和重复表达，需要润色一下）
Crucially, unlike short-horizon or loosely coupled GUI tasks, tasks in \BenchName{} are deliberately constructed and annotated to exhibit substantial cross-step causal dependencies.
Later actions often depend on intermediate outcomes produced by earlier steps, which we refer to as \emph{intermediate task states}.
These intermediate states are neither directly observable from the initial instruction nor recoverable solely from the current GUI state.
As a result, executing tasks in \BenchName{} requires agents to correctly establish and maintain task-relevant intermediate states over extended interaction horizons, therefore making the effective utilization of interaction history and intermediate states a prerequisite for task success, rather than an optional enhancement. To make such intermediate task states explicit and amenable to analysis, we annotate sparse \emph{\StateAnchors{}} along the trajectory, each summarizing a task-relevant state change or intermediate outcome that constrains subsequent steps. 

\noindent \textit{Task Types.}
We observe that in multi-app environments, tasks cannot be adequately characterized using coarse scenario labels such as shopping or travel alone.
Even when involving similar combinations of applications, tasks may exhibit substantially different interaction patterns, step dependencies, and behavior distributions, largely due to differences in user intent.

% To capture this distinction, we introduce the notion of \emph{primary intent}, which describes the core objective driving task execution.
% Motivated by real-world application usage, where interactions are typically organized around a single dominant goal, we assume that each task in \BenchName{} is associated with one primary intent.
% Based on this formulation, we categorize tasks into the following eight intent classes:
To capture this distinction, we define \emph{primary intent} as the dominant objective that drives task execution.
Following real-world app usage, where user interactions are typically organized around a single goal, we assign each task in \BenchName{} exactly one primary intent.
Accordingly, we group tasks into 8 intent classes:
(1) Lookup;
(2) Compare \& Decide;
(3) Purchase / Order;
(4) Booking / Reservation;
(5) Communicate / Coordinate;
(6) Share / Recommend;
(7) Create Content;
and (8) Configure / Authorize.
See Appendix~\ref{app:androtrace_task_types} for details.
% 动作空间设计
% \textbf{Action Space}:open_app,tap,long_press,swipe,swipe_tow_points,wait,finish,home,back. 其中，swipe表示根据起始点和方向滑动，而swipe_two_points表示根据起始点和终点滑动，测评过程中，swipe和swipe_two_points操作被简化为滑动方向是否匹配，详情见表
% 等宽字体
% \paragraph{Action Space.} 
% The action space of \BenchName{} includes 11 types of actions: 
% \texttt{OPEN\_APP}, \texttt{TAP}, \texttt{LONG\_PRESS}, \texttt{SWIPE}, \texttt{INPUT\_TEXT}, 
% \texttt{SWIPE\_TWO\_POINTS}, \texttt{WAIT}, \texttt{CAPTURE\_SCREEN},  \texttt{HOME}, \texttt{BACK}, and \texttt{FINISH}. 

\noindent \textit{Action Space.}
The action space of \BenchName{} includes 11 action types:
\act{open\_app}, \act{tap}, \act{long\_press}, \act{swipe}, \act{input\_text},
\act{swipe\_two\_points}, \act{wait}, \act{capture\_screen}, \act{home}, \act{back}, \act{finish}.
\act{swipe} represents sliding based on the starting point and direction, while \act{swipe\_two\_points} represents sliding based on the starting point and ending point. During the evaluation, \act{swipe} and \act{swipe\_two\_points} operations were simplified to whether the sliding directions matched. See Appendix~\ref{app:androtrace_app_set} for details.

\subsection{Data Pipeline}
\label{sec:3_data_pipeline}

\begin{figure}[t]
    \centering
    \includegraphics[width=0.95\linewidth]{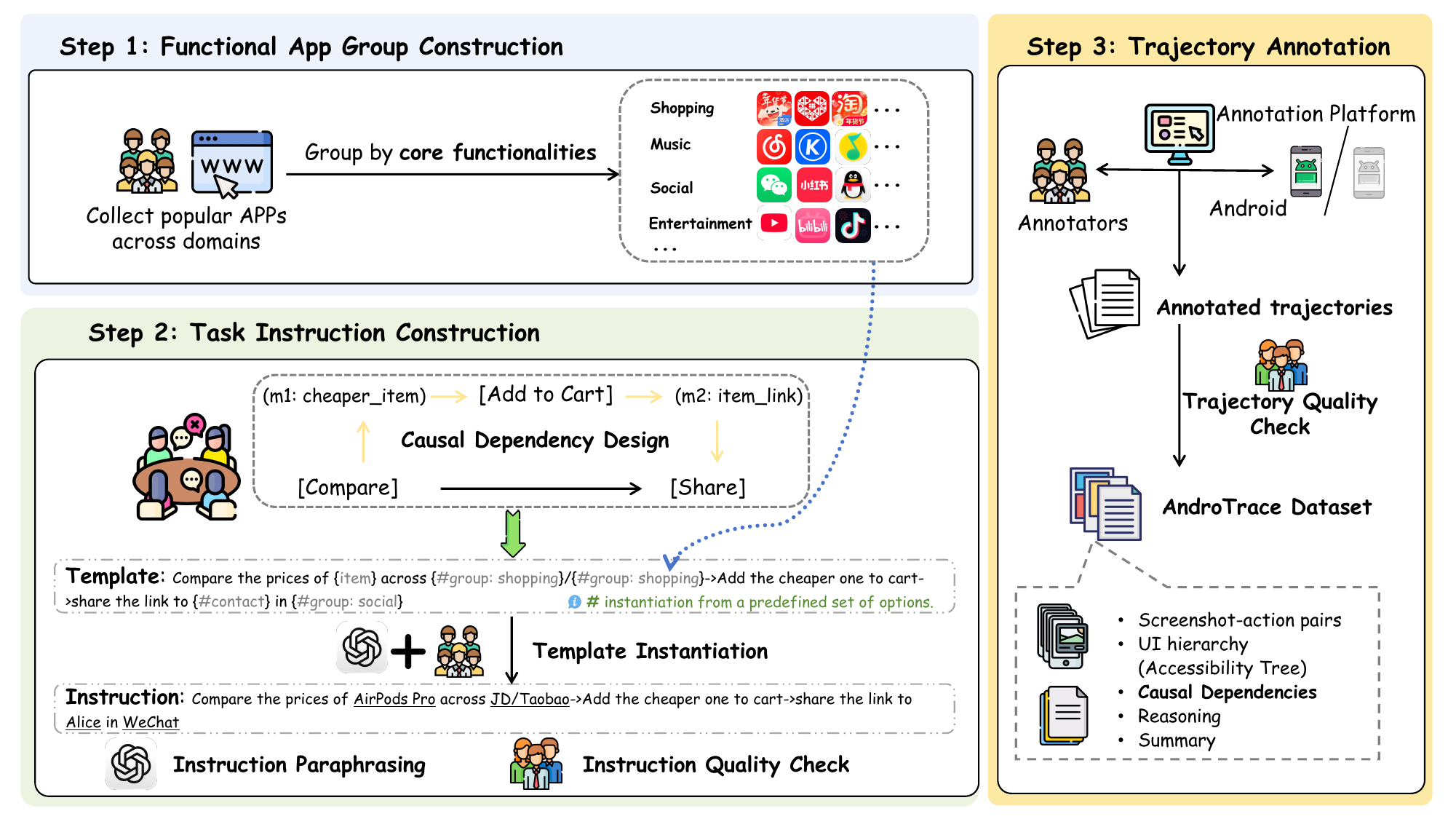}
    \caption{
    % Overview of the \BenchName{} dataset construction pipeline. (Step 1) We collect popular mobile apps and organize them into functional app groups. (Step 2) We construct long-horizon cross-app task instructions with explicit step-to-step causal dependencies via dependency-aware templates and instantiation. (Step 3) Annotators execute tasks on Android devices or emulators via an annotation platform, producing annotated trajectories and quality-checked dataset outputs. \\
    Overview of the AndroTMem-Bench dataset construction pipeline. (1) Collect popular mobile apps and group them by function. (2) Generate long-horizon cross-app task instructions with step-to-step causal dependencies using dependency-aware templates. (3) Execute and annotate tasks on Android devices or emulators, producing quality-checked trajectories and dataset outputs.
    }
    \label{fig:3_dataset_pipeline}
    % \vspace{-2em}
\end{figure}

% 半自动 OR 全自动？
% 鉴于 \longTask 的任务复杂性，尤其是其长程依赖与细粒度中间状态的需求，现有 LLM 难以在不引入额外人工校正的情况下生成高质量任务（即便由 LLM 生成的任务描述，仍需 GUI 领域专家进行修订）。因此，完全自动化的数据构建流程——即在任务设计、数据采集与后处理环节均不需要人工参与——并不适用于我们的设定。另一方面，从既有数据中“清洗”得到所需样本通常要求对完整交互轨迹进行逐步处理，以提取关键中间状态，这对人工成本与自动化处理能力都提出了较高要求。基于上述考虑，我们采用半自动标注范式：由具备丰富 GUI 经验的专家负责任务设计，并完成数据标注工作。
% Specifically，First，任务设计层面，
An overview of the entire data pipeline is illustrated in Figure~\ref{fig:3_dataset_pipeline}, which consists of task instruction construction (Step~1--2) and trajectory annotation (Step~3).

% \paragraph{Motivation for a Semi-Automatic Pipeline.}
% Given the complexity of \LongTask, especially the cross-step causal dependencies and the need for fine-grained intermediate states, current LLMs are insufficient to meet our requirements. In practice, tasks generated by LLMs still require substantial corrections from GUI experts, which makes a fully automatic pipeline, where task design, data collection, and post-processing proceed without human involvement—inappropriate for our setting. Moreover, constructing our desired annotations by re-annotating existing datasets typically requires processing entire interaction trajectories to recover intermediate states, which imposes high demands on both human labor and automation capabilities. Therefore, we adopt a semi-automatic annotation paradigm, where experienced GUI experts design the tasks and perform the annotations. 
\noindent \textit{Motivation for a Semi-Automatic Pipeline.}
\LongTask{} requires long-horizon execution with explicit \Dependency{} and fine-grained intermediate states, which we record as sparse \emph{\StateAnchors{}} that must be correct and consistent across the entire trajectory.

While fully automatic pipelines can scale data collection, they typically optimize for obtaining raw interaction traces, and thus provide limited control over (i) intent-faithful task design, (ii) the correctness of intermediate causal states, and (iii) the consistency of dependency links across steps and apps.
In our preliminary trials, LLM-generated tasks and auto-collected trajectories frequently exhibit mismatches between the intended goal and the executed behavior, and recovering the required intermediate states afterward would still require processing the full trajectory—either by extensive human re-annotation or by automation that is not yet reliable under long causal chains.

We therefore adopt a semi-automatic pipeline that places human expertise only where it is indispensable:
experienced GUI experts specify intent-driven tasks, while the platform automates closed-loop action execution with real-time visual feedback,
synchronized collection of UI states,
and the management of configurable structured annotation fields.
Compared to purely automatic data collection, this design yields annotations that are (a) causally grounded, (b) fine-grained and directly usable for evaluation, and (c) significantly cheaper to produce than full post-hoc re-annotation of long trajectories.

% 任务Instruction构建
% APP功能组构建-功能组特定组合产生任务模板（参数插槽）-任务生成-人工调整-同义转换
% 任务分类：主目标意图
% 构建带有明确中间状态的任务
% 需要说明的是，不少于20步并非\LongTask的本质定义条件，而是一个经验性的选择，目的
% 我们首先收集了一组常用的移动应用程序，并根据其核心功能将它们组织成不同的应用程序组（\textbf{\APPGroup{}}），同一组内的应用程序具有相似的交互模式和功能语义。基于这些分组，我们设计了 70 个以上的跨应用任务模板，这些模板经过精心构建，旨在在应用程序之间建立强烈的步骤依赖关系。
% 每个任务模板都包含应用程序功能组和任务参数的占位符（例如，搜索查询、目标联系人、消息内容等）。重要的是，这些参数仅在模板级别进行实例化，并且不会完全指定后续步骤所需的中间结果。因此，成功执行任务需要智能体正确地生成、跟踪和跨应用程序重用中间任务状态，而不是简单地遵循一系列松散耦合的指令。对于特定应用程序独有的功能，我们会明确指定相应的应用程序以保持真实性。完成模板填充后，我们进一步使用 GPT-4o 对生成的指令进行语言改写，在保留关键信息与任务约束的前提下，将其转换为更加自然、贴近人类表达习惯的自然语言指令。
% \paragraph{Task Instruction Construction.} 
\noindent \textit{Task Instruction Construction.}
We first collect about 50 commonly used mobile applications and organize them into distinct application groups(\emph{\APPGroup{}}. See Appendix~\ref{app:androtrace_app_categories} for details.) based on their core functionalities, where apps within the same group share similar interaction patterns and functional semantics. Based on these groups, we design over 70 cross-app task templates that are deliberately constructed to induce strong step-to-step dependencies across applications. 
Each task template contains slots for application function groups and task parameters (e.g., search queries, target contacts, or message contents). These parameters are instantiated only at the template level and do not fully specify the intermediate results required by subsequent steps. For functionalities that are unique to a specific application, the corresponding app is explicitly specified to preserve realism. After instantiation of the template, we further employ GPT-4o to rewrite the generated instructions, transforming them into more natural, human-like expressions while preserving the essential task information and constraints.

\noindent \textit{Data Annotation.}
In the annotation process, the annotator first acquires the screenshot of the current step through the platform, with the accessibility tree synchronized simultaneously. The platform allows annotators to inspect the coordinates and other properties of target UI elements in real time and to annotate the corresponding action accordingly.  The annotator then records the \StateAnchors{}, reasoning analysis, and operation summary for the current step, and executes the annotated action via the platform to transition the device to the next state. This process is repeated until the task is completed, which is marked by a \act{finish} action.

% fig_qualitative.tex
\begin{figure*}[htbp]
    \centering

    % 第一行
    \includegraphics[width=0.28\textwidth]{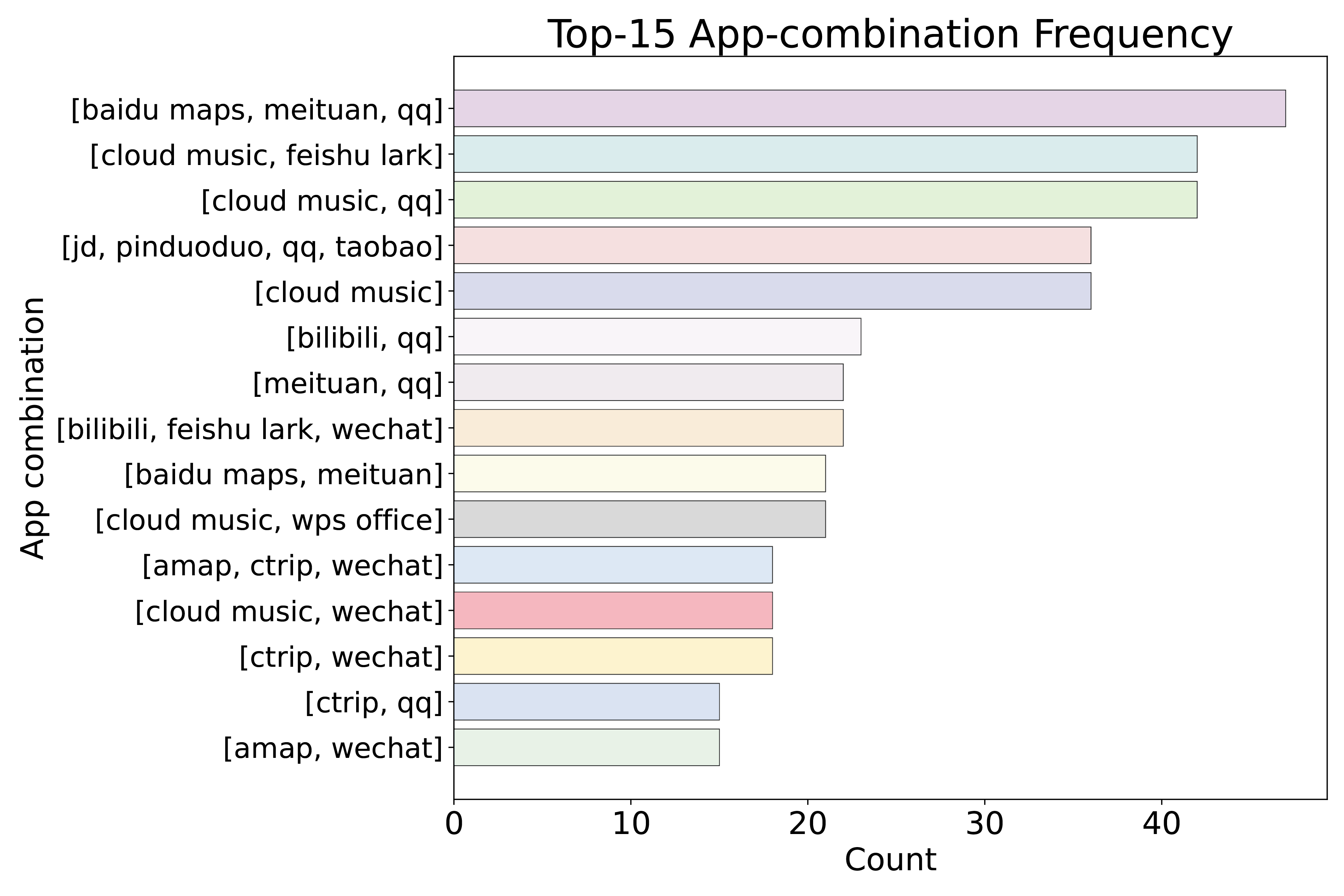}\hspace{2mm}
    \includegraphics[width=0.28\textwidth]{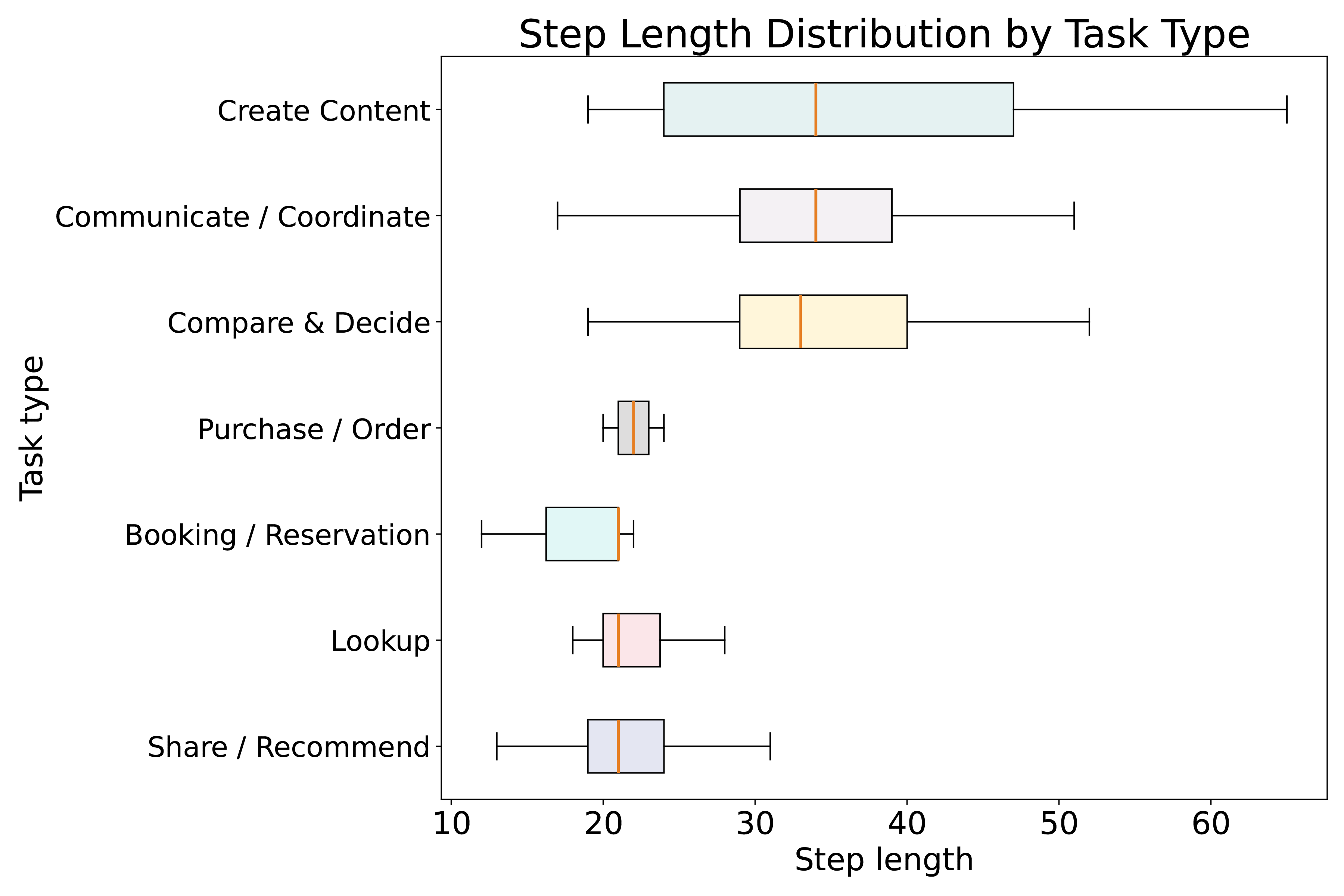}\hspace{2mm}
    \includegraphics[width=0.28\textwidth]{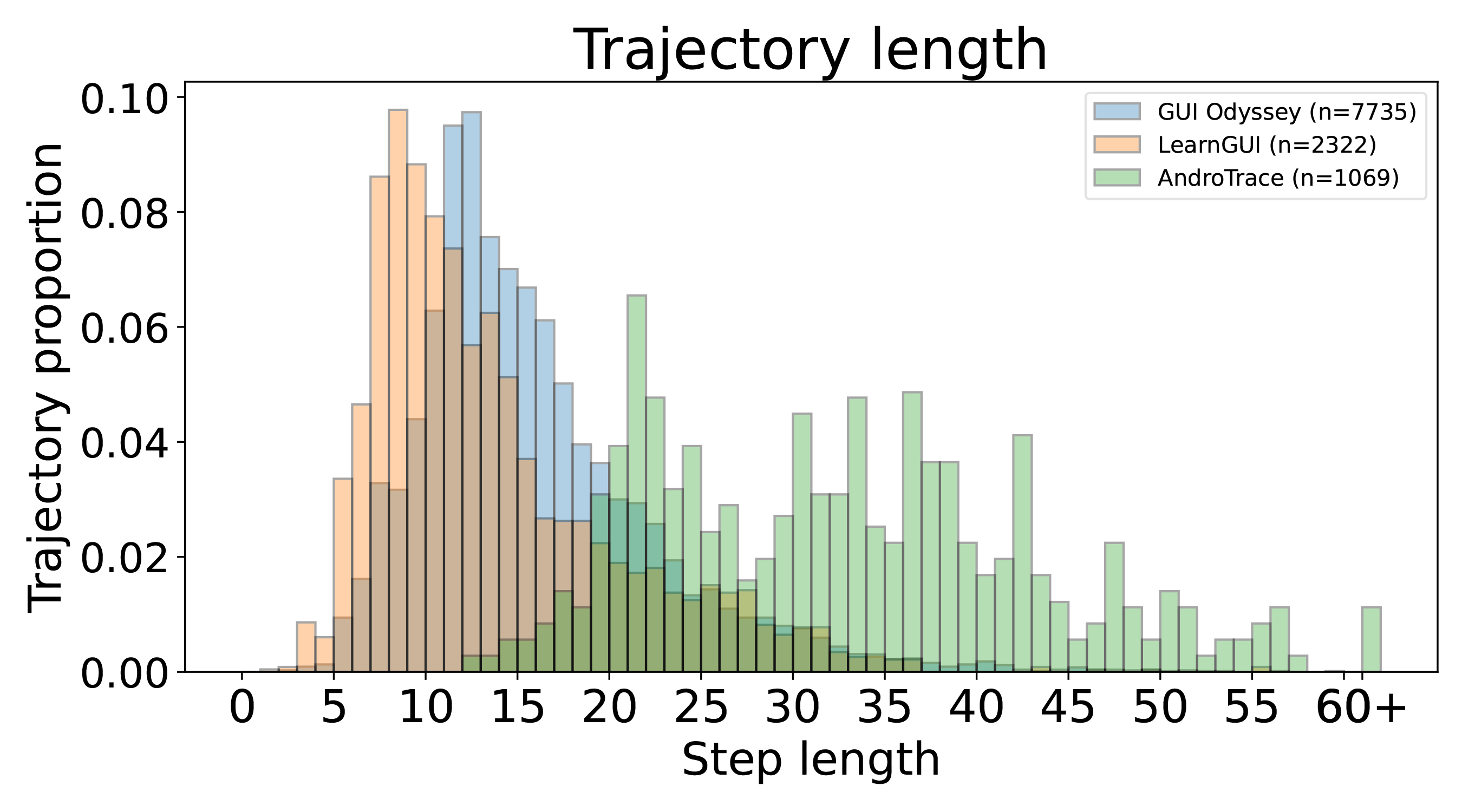}

    \vspace{1mm}

    % 第二行
    \includegraphics[width=0.28\textwidth]{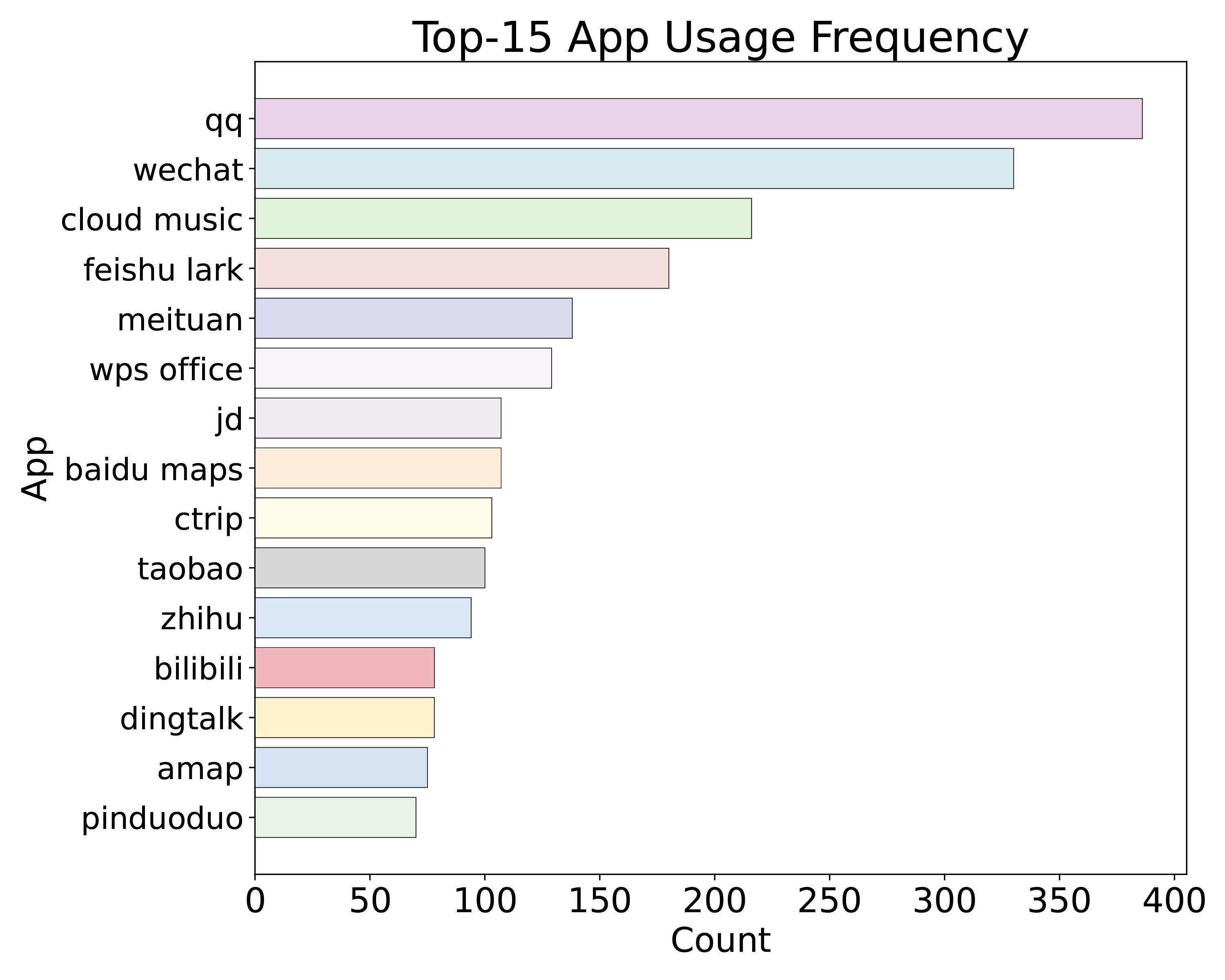}\hspace{2mm}
    \includegraphics[width=0.28\textwidth]{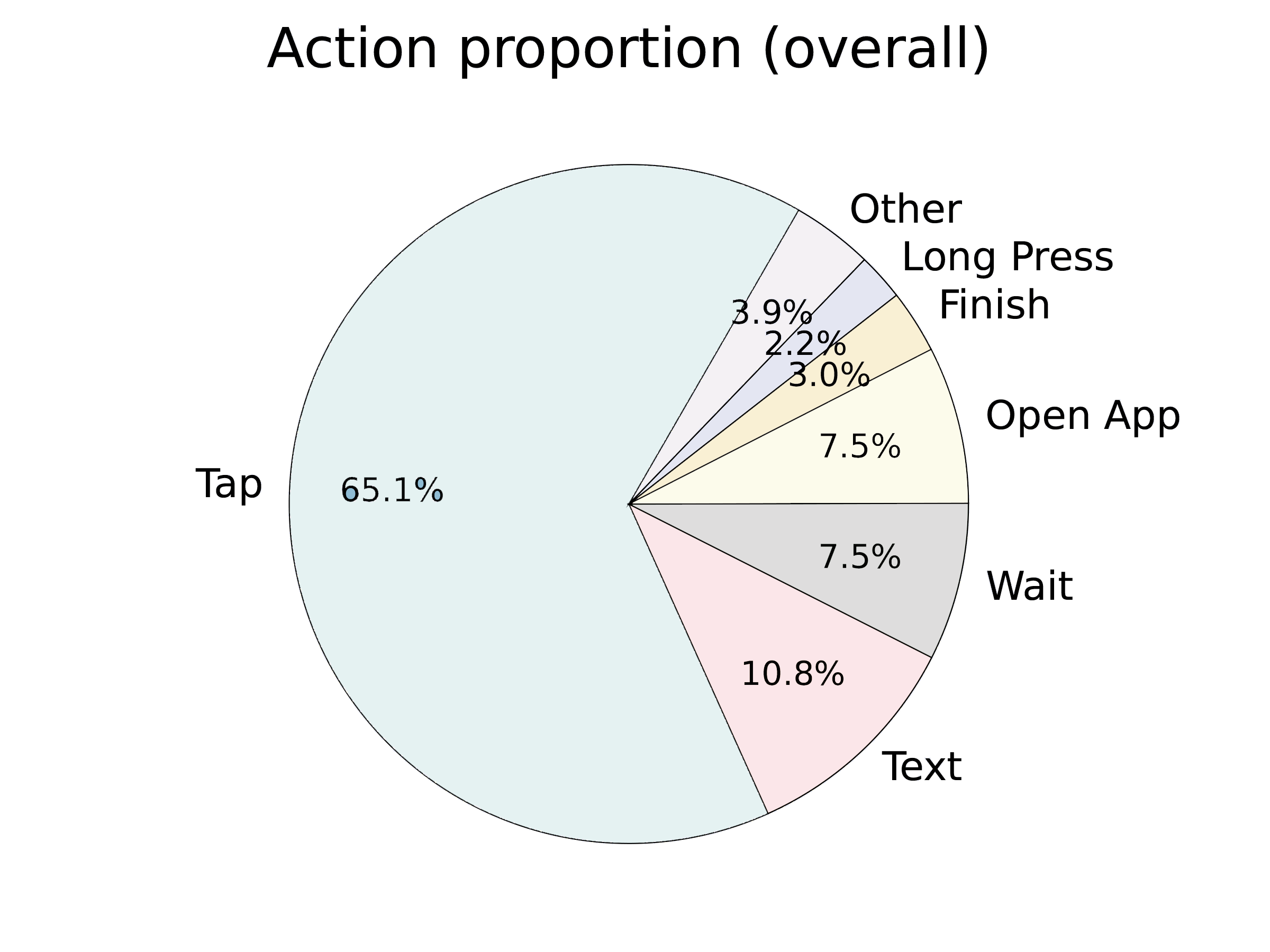}\hspace{2mm}
    \includegraphics[width=0.28\textwidth]{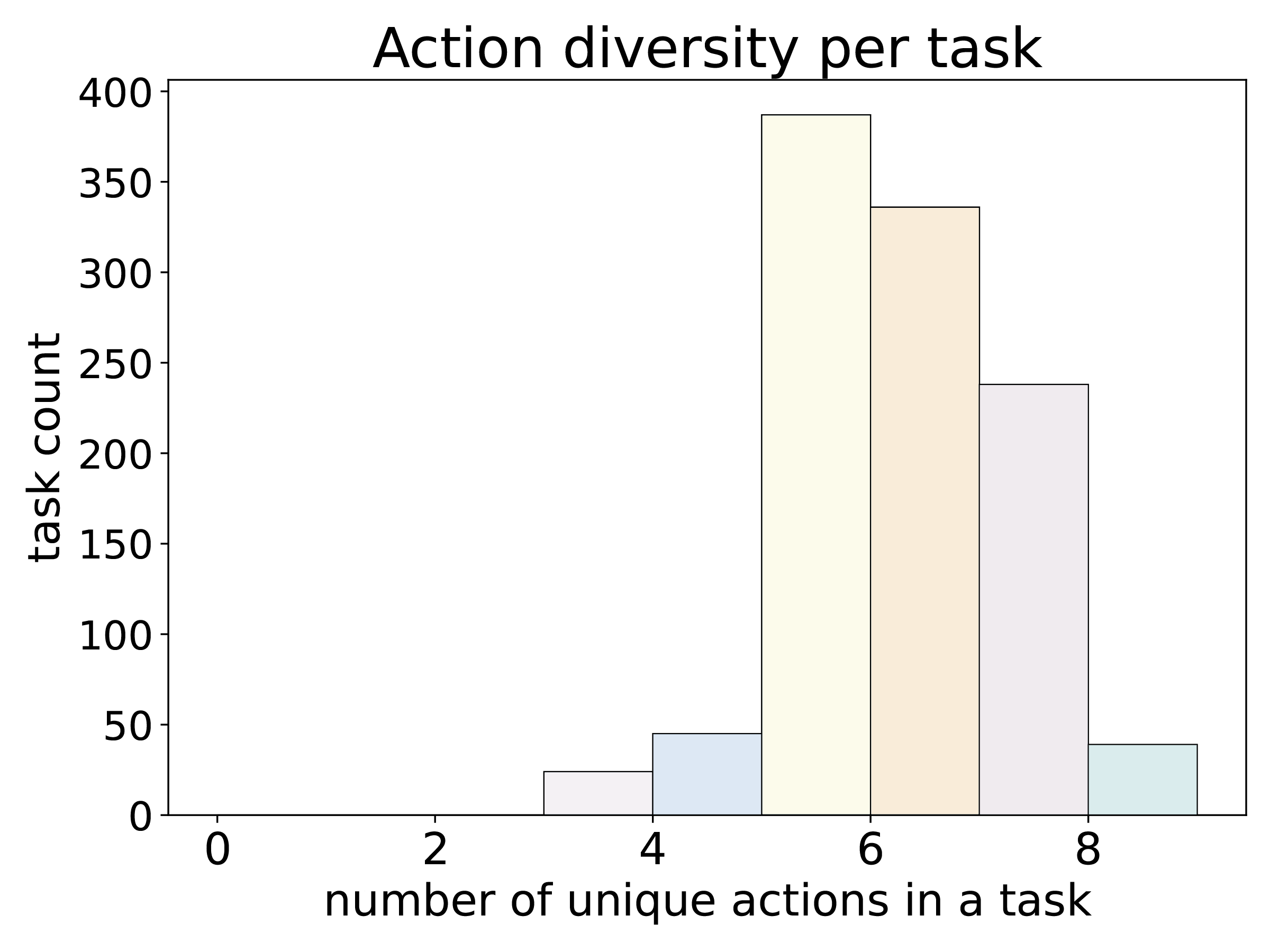}

    \caption{
Overview statistics of AndroTMem-Bench. The first row reports the top app combinations, step length distribution by task type, and overall trajectory length distribution (with comparison to prior benchmarks). The second row shows app usage frequency, overall action-type proportions, and action diversity per task.
    }
    \label{fig:3_dataset_static}
    % \vspace{-2em}
\end{figure*}

% 数据质量检查主要包括两部分：指令检查与轨迹检查。指令检查在标注开始前进行，由reviewers对任务指令进行审核，判断其是否存在表述不清或歧义之处；如有问题，则对指令进行编辑、修订或筛除。轨迹检查则针对标注者产出的交互轨迹，检查标准包括：每一步是否配备对应的屏幕截图；任务指令与实际执行轨迹是否一致；以及额外标注信息（如 Summary、Reasoning 等）是否与对应步骤内容相匹配。经过上述检查与整理，我们最终构建了数据集 \BenchName。
% \paragraph{Data Quality Control.} 
\noindent \textit{Data Quality Control.}
Our data quality control consists of two components: \textbf{instruction validation} and \textbf{trajectory validation}. Instruction validation is performed before annotation begins: designated reviewers audit each task instruction to identify unclear or ambiguous descriptions, and then edits, revises, or filters out problematic instructions when necessary. Trajectory validation is applied to the annotated interaction traces, with criteria including: (i) whether each step is associated with a corresponding screenshot, (ii) whether the executed trajectory matches the task instruction, and (iii) whether extra annotations (e.g., Summary and Reasoning) are aligned with the corresponding steps. After this quality checking and consolidation process, we obtain the dataset \BenchName.

\subsection{Data Statics}
\label{sec:3_data_statics}

% 我们的数据集共包含 1069 个高质量 GUI 任务指令，对应 34,473 个交互步骤，平均每个任务包含 32.1 个步骤。据我们所知，这是目前安卓移动端公开 GUI 数据集中任务行程最长的数据集之一。数据集覆盖 50 个常用移动应用，涵盖多种应用功能与跨应用交互场景。与现有数据集不同的是，我们引入了 里程碑锚点（Milestone Anchors） 机制，对任务执行过程中的关键中间状态进行显式建模，从而支持对历史信息利用与状态管理能力的系统评估。
Our dataset consists of 1,069 high-quality GUI tasks with substantial step-to-step causal dependencies, comprising 34,473 interaction steps in total, with an average of 32.1 steps per task and a maximum of 65 steps.
% To the best of our knowledge, this represents one of the longest-horizon GUI datasets for Android mobile devices to date. 
The dataset spans 50 widely used mobile applications, covering diverse functionalities and cross-app interaction scenarios. In addition to raw trajectories, \BenchName{} provides step-aligned auxiliary annotations, including reasoning traces, summaries, and sparse \emph{state anchors} that mark task-relevant intermediate states for analysis and evaluation.
Figure ~\ref{fig:3_dataset_static} presents key statistical properties of our dataset.

\section{AndroTMem-Bench}
\label{sec:5_experiment}

\subsection{Experiment Setup}
\label{sec:5_setup}
\noindent \textit{Setup.}
We evaluate existing GUI agents on \textbf{\BenchName{}-Bench} to characterize the intrinsic challenges posed by long-horizon, causally dependent GUI tasks. 
% All agents are evaluated on the same set of tasks using their default interaction pipelines, without modifying their internal history modeling mechanisms. 
This evaluation measures the out-of-the-box capability of current GUI agents to solve long-horizon, cross-app tasks with strong step-to-step dependencies. Additional implementation and evaluation details are provided in Appendix~\ref{app:experiment}.

\noindent \textit{Metrics.}
For step-level evaluation, we adopt \textbf{Action Matching Score (AMS)} following GUI-Odyssey and AITW. 
An action is considered correct if its type matches the ground truth. 
For \act{tap} and \act{long\_press}, predictions are correct when the normalized screen distance to the annotated point is within 14\% or the predicted point falls inside the target UI element segmented by SAM2. 
For \act{swipe} actions, correctness is determined by direction, and for \act{text}, we use the Average Normalized Levenshtein Similarity (ANLS).

To measure long-horizon interaction memory capability, we introduce \textbf{Task Completion Rate (TCR)} based on \textit{state anchors}. 
A task is considered successful if the agent reaches the final anchor while satisfying the causal dependencies among preceding anchors, reflecting whether task-critical intermediate states are correctly preserved and reused.

We further report two efficiency metrics: \textbf{Avg. Tokens}, the average number of tokens consumed per interaction step, and \textbf{Avg. Time}, the average execution time per step. 
These metrics characterize the computational cost of different history utilization strategies during long-horizon execution.

% 当历史信息以不同结构出现时，GUI Agent 的能力上限是不同的
% 为了隔离历史表示的影响，我们让不同的模型在相同规则下生成其自己的摘要或锚点。

% \subsection{Main Results}
% \label{sec:5_results}
% % Main Results：讲“事实 + 现象 + 排序”，不讲“为什么”
% % “What happens on your benchmark?”
% \paragraph{Overall performance.}
\subsection{Benchmark Results and Diagnosis}
\label{sec:4_results_analysis_bench}
% \paragraph{Overall Benchmark Performance.}
\noindent \textit{Overall Benchmark Performance.}
Table~\ref{tab:5_model_performance} reports the performance of all evaluated agents on \BenchName{}-Bench, including both closed-source and open-source systems as well as multi-agent frameworks.
Overall, all agents exhibit relatively low absolute performance, indicating that long-horizon Android GUI tasks remain highly challenging for current approaches.
Among closed-source models, Gemini-3-Flash achieves the strongest performance with an AMS of 46.14\% and a TCR of 55.21\%.
However, most other models remain far below this level.
For instance, GPT-4o and GPT-5 achieve AMS scores of only 14.24\% and 12.37\% respectively, with similarly low TCR values.
A similar pattern is observed among open-source agents, where UI-TARS-1.5-7B performs best but still achieves only 35.62\% AMS and 34.55\% TCR.
These results suggest that even the strongest existing GUI agents are far from reliably solving long-horizon cross-app workflows. Maintaining consistent task execution across extended interaction trajectories remains an open challenge.
\setlength{\tabcolsep}{5pt}
\renewcommand{\arraystretch}{1.15}

\begin{wraptable}{r}{0.47\columnwidth}

\centering
\caption{Overall performance of evaluated GUI agents on \BenchName{}-Bench. AMS measures step-level action accuracy, while TCR evaluates task completion based on anchor-defined intermediate states.}
\label{tab:5_model_performance}

\resizebox{0.85\linewidth}{!}{%
\begin{tabular}{lcc}
\toprule
\textbf{Model} & \textbf{AMS} & \textbf{TCR} \\
\midrule
\multicolumn{3}{c}{\textbf{Closed-Source}} \\
% \cmidrule(lr){1-3}
\midrule
GPT-4o~\cite{hurst2024gpt} & 14.24 & 11.75 \\
GPT-5~\cite{singh2025openai} & 12.37 & 11.46 \\
Gemini-2.5-Flash~\cite{comanici2025gemini} & 15.62 & 14.51 \\
Gemini-2.5-Pro~\cite{comanici2025gemini} & 32.71 & \underline{41.11} \\
Gemini-3-Flash & \textbf{46.14} & \textbf{55.21} \\

\midrule
\multicolumn{3}{c}{\textbf{Open-Source}} \\
% \cmidrule(lr){1-3}
\midrule
Qwen2.5-VL-7B~\cite{bai2025qwen2} & 20.71 & 16.04 \\
UI-TARS-1.5-7B~\cite{qin2025ui} & \underline{35.62} & 34.55 \\
AgentCPM-GUI-8B~\cite{zhang2025agentcpm} & 14.06 & 10.49 \\
UI-Venus-Navi-7B~\cite{gu2025ui} & 14.08 & 9.64 \\
InfiGUI-R1~\cite{liu2025infigui} & 14.12 & 9.20 \\

\midrule
\multicolumn{3}{c}{\textbf{Multi-Agent}} \\
% \cmidrule(lr){1-3}
\midrule
Mobile-Agent-E~\cite{wang2025mobile} & 15.77 & 11.71 \\
COLA~\cite{zhao2025cola} & 17.22 & 11.85 \\

\bottomrule
\end{tabular}
}
\end{wraptable}

% % \documentclass{article}
% % \usepackage{booktabs}
% % \usepackage{geometry}
% % \usepackage{array}
% % \usepackage{multirow}
% % \geometry{a4paper, margin=0.6in} % 充足页边距

% % % 仅把行高从1.4调为1.05，精准压缩多余间距，其他不变
% % \setlength{\tabcolsep}{12pt} % 保留原列间距
% % \renewcommand{\arraystretch}{1.05} % 核心修改：行高微缩，无多余间距

% % \begin{document}

% \begin{table*}[!t]
% \centering
% \caption{Model Performance Comparison (AMS \& SR)}
% % 完全保留原列格式，3cm模型列左对齐+6列1.5cm居中
% \begin{tabular}{>{\raggedright\arraybackslash}m{3cm} *{6}{>{\centering\arraybackslash}m{1.5cm}}}
% \toprule
% % 保留Model跨两行+\vspace{2pt}下移，完全不变
% \multirow{2}{*}{\textbf{\makecell{\vspace{2pt}Model}}} & \multicolumn{2}{c}{\textbf{RAW}} & \multicolumn{2}{c}{\textbf{Summary}} & \multicolumn{2}{c}{\textbf{AR}} \\
% \cmidrule(lr){2-3} \cmidrule(lr){4-5} \cmidrule(lr){6-7}
% & AMS & SR & AMS & SR & AMS & SR \\
% \midrule
% % 闭源分类行，完全保留原格式，无额外修改
% \multicolumn{7}{>{\centering\arraybackslash}m{13.5cm}}{\textbf{Close-Source}} \\
% \cline{1-7}
% GPT-4o    & 14.24 & 11.75 & 15.49 & 13.02 & 19.83 & 17.72 \\
% \midrule
% % 开源分类行，完全保留原格式，无额外修改
% \multicolumn{7}{>{\centering\arraybackslash}m{13.5cm}}{\textbf{Open-Source}} \\
% \cline{1-7}
% Qwen2.5-VL-7B      & 20.71 & 16.04 & 35.09 & 42.39 & 40.36 & 46.19 \\
% UI-TARS-1.5-7B   & 35.62 & 34.55 & 44.49 & 43.84 & 46.07 & 46.21 \\
% AgentCPM-GUI-8B       & 14.06 & 10.49 & 21.40 & 22.96 & 28.63 & 27.89 \\
% \bottomrule
% \end{tabular}
% \label{tab:model_performance}
% \end{table*}

% % \end{document}

% \paragraph{Performance across Task Types.}
\noindent \textit{Performance across Task Types.}
% defined in Table~\ref{tab:5_model_performance_by_group_row}.
To better understand where current agents struggle, we further analyze performance across different user intent categories.
% These task types reflect common mobile interaction goals such as information lookup, decision making, purchasing, communication, and content creation.
Table~\ref{tab:5_model_performance_by_group_row} reports the performance of GUI agents across seven representative intent categories.
% 从不同任务类型的难度不同到对不同agent的区分
% We observe that tasks requiring cross-step information reuse and cross-app coordination tend to be significantly more difficult.
% For example, tasks under \textit{Compare \& Decide} and \textit{Purchase/Order} require agents to retrieve information from multiple sources and make downstream decisions based on earlier observations.
We observe substantial performance variation across intents. In particular, intents that involve non-local state reuse or coordination across apps are often among the most challenging for many agents.
For example, tasks under \textit{Compare \& Decide} often require agents to gather and reconcile information from multiple sources, while \textit{Purchase/Order} tasks frequently depend on carrying intermediate state across multiple steps toward successful completion.
% Similarly, \textit{Communicate} and \textit{Share / Recommend} tasks often require extracting information from one application and transferring it to another.
These workflows involve long dependency chains where later actions critically depend on intermediate results obtained several steps earlier. 
In practice, such dependency-heavy intents are often more challenging for many agents than more localized interaction patterns, although the exact difficulty varies across models. 
This suggests that a key challenge in these scenarios lies not only in individual action prediction, but also in the ability to preserve and reuse task-critical intermediate information across the interaction trajectory.

\newcommand{\best}[1]{\textbf{#1}}
\newcommand{\second}[1]{\underline{#1}}

\begin{table*}[t]
\centering
\caption{Per-intent AMS and TCR of evaluated agents on seven user intents in AndroTMem-Bench.
Intent categories follow the taxonomy in Table~\ref{tab:task_types_daily}.
Abbreviations denote task types (e.g., Lkp: Lookup, CpD: Compare \& Decide, Pur: Purchase / Order).}
\label{tab:5_model_performance_by_group_row}

\setlength{\tabcolsep}{3pt}
\renewcommand{\arraystretch}{1.10}
\footnotesize

\begin{adjustbox}{width=\linewidth}
\begin{tabular}{lcccccccccccccc}
\toprule
\textbf{Model} &
\multicolumn{2}{c}{\textbf{CC}} &
\multicolumn{2}{c}{\textbf{Cre}} &
\multicolumn{2}{c}{\textbf{CpD}} &
\multicolumn{2}{c}{\textbf{SeR}} &
\multicolumn{2}{c}{\textbf{Lkp}} &
\multicolumn{2}{c}{\textbf{Bok}} &
\multicolumn{2}{c}{\textbf{Pur}} \\
\cmidrule(lr){2-3}\cmidrule(lr){4-5}\cmidrule(lr){6-7}\cmidrule(lr){8-9}
\cmidrule(lr){10-11}\cmidrule(lr){12-13}\cmidrule(lr){14-15}
& \scriptsize AMS & \scriptsize TCR
& \scriptsize AMS & \scriptsize TCR
& \scriptsize AMS & \scriptsize TCR
& \scriptsize AMS & \scriptsize TCR
& \scriptsize AMS & \scriptsize TCR
& \scriptsize AMS & \scriptsize TCR
& \scriptsize AMS & \scriptsize TCR \\
\midrule

% -------------------- Closed-Source --------------------
\multicolumn{15}{c}{\textbf{Closed-Source}} \\
% \cmidrule(l{10pt}r{10pt}){1-15}
\midrule
\scriptsize GPT-4o
& 18.72 & 18.08
& 22.07 & 18.95
& 20.40 & 18.64
& 19.10 & 13.45
& 21.23 & 11.90
& 20.75 & 10.21
& 18.18 & 9.35 \\
\scriptsize GPT-5
& 12.35 & 13.44
& 12.15 & 11.66
& 10.95 & 11.40
& 16.05 & 14.62
& 10.61 & 7.14
& 24.53 & 20.15
& 11.36 & 9.38 \\
\scriptsize Gemini-2.5-Flash
& 25.21 & 33.20
& 27.93 & 35.86
& 29.86 & 32.68
& 29.05 & 34.50
& 17.88 & 16.67
& 35.85 & 30.00
& 34.09 & 22.22 \\
\scriptsize Gemini-2.5-Pro
& 30.81 & 39.67
& 33.99 & 47.80
& 32.09 & 36.67
& 43.79 & 43.75
& 37.50 & \second{47.06}
& 35.15 & 12.50
& 33.45 & 41.07 \\
\scriptsize Gemini-3-Flash
& \second{46.21} & \textbf{56.13}
& \textbf{48.12} & \textbf{64.72}
& \second{42.63} & \second{46.05}
& \second{50.24} & \textbf{56.73}
& \second{43.58} & \textbf{52.38}
& \textbf{43.40} & \second{50.00}
& 40.91 & 44.44 \\
\midrule

% -------------------- Open-Source --------------------
\multicolumn{15}{c}{\textbf{Open-Source}} \\
% \cmidrule(l{10pt}r{10pt}){1-15}
\midrule
\scriptsize Qwen2.5-VL-7B
& 41.71 & \second{46.11}
& \second{41.84} & \second{53.80}
& 40.20 & 43.13
& 34.18 & 41.08
& 28.76 & 36.81
& 32.34 & 28.57
& \second{43.18} & 44.44 \\
\scriptsize UI-Venus-Navi-7B
& 16.88 & 16.11
& 20.36 & 14.58
& 21.45 & 12.72
& 21.35 & 11.70
& 23.46 & 9.52
& 16.98 & 10.76
& 27.27 & 33.33 \\
\scriptsize UI-TARS-1.5-7B
& \textbf{46.94} & 44.94
& 40.90 & 47.93
& \textbf{47.86} & \textbf{48.23}
& \textbf{50.73} & \second{46.45}
& \textbf{44.07} & 38.71
& \second{43.23} & \textbf{53.06}
& \textbf{52.27} & \textbf{55.56} \\
\scriptsize AgentCPM-GUI-8B
& 25.50 & 25.15
& 35.41 & 40.87
& 30.44 & 23.70
& 26.97 & 25.85
& 30.13 & 32.26
& 19.80 & 26.53
& 27.27 & 11.11 \\
\scriptsize InfiGUI-R1
& 12.50 & 8.26
& 15.83 & 10.42
& 14.22 & 8.89
& 13.33 & 8.33
& 21.40 & 14.28
& 18.66 & 33.33
& 22.80 & \second{50.00} \\
\midrule

% -------------------- Multi-Agent --------------------
\multicolumn{15}{c}{\textbf{Multi-Agent}} \\
% \cmidrule(l{10pt}r{10pt}){1-15}
\midrule
\scriptsize COLA
& 15.42 & 10.92
& 17.26 & 12.45
& 18.05 & 12.01
& 16.88 & 14.28
& 16.43 & 9.85
& 15.82 & 7.92
& 12.58 & 12.38 \\
\scriptsize Mobile-Agent-E
& 16.50 & 10.50
& 16.00 & 12.00
& 17.46 & 13.50
& 21.47 & 14.50
& 18.21 & 8.61
& 24.79 & 15.00
& 22.23 & 18.54 \\

\bottomrule
\end{tabular}
\end{adjustbox}
\end{table*}
% \paragraph{Diagnosis: Memory as the Primary Bottleneck.}

\noindent \textit{Diagnosis: Memory as the Primary Bottleneck.}
We further analyze agent performance with respect to interaction horizon length.
As shown in Figure~\ref{fig:1_teaser}, performance consistently degrades as the number of steps increases across all evaluated models.
This decline cannot be explained solely by local perception or action prediction errors.
Instead, many failures occur when agents must recall and reuse information obtained earlier in the interaction, such as retrieved prices, selected items, or identified contacts.
This pattern aligns with the design of \BenchName{}-Bench, where sparse but causally critical intermediate states determine downstream decisions.
However, existing history representations—whether raw trajectories or coarse summaries—often fail to preserve these key states as trajectories grow longer.
These results indicate that the primary bottleneck in long-horizon GUI interaction lies in representing and retrieving task-critical intermediate states across the interaction trajectory.
% \paragraph{Implication for Interaction Memory Design.}

\noindent \textit{Implication for Interaction Memory Design.} The above diagnosis indicates that effective long-horizon GUI interaction requires more than accurate perception or action prediction.
Agents must be able to explicitly represent and reuse task-critical intermediate states that form the causal backbone of the interaction trajectory.
Motivated by this observation, we introduce ASM, a causally structured interaction memory mechanism that organizes past interactions around state anchors and their causal dependency relations.
By explicitly modeling intermediate states that determine downstream decisions, ASM enables agents to retrieve the most relevant task information without relying on excessively long raw histories.
In the following section, we describe the design of ASM and evaluate how structured anchor-based memory affects agent performance on \BenchName{}-Bench.

\section{\methodname{}}
\label{sec:4_methdology}
\subsection{Motivation from Benchmark Diagnosis}
\label{sec:6_motivation_methdology}
\begin{figure*}[t]
    \centering
    \captionsetup[subfigure]{justification=centering}
    % Row 1
      \begin{subfigure}[t]{0.32\linewidth}
        \centering
        \includegraphics[width=\linewidth]{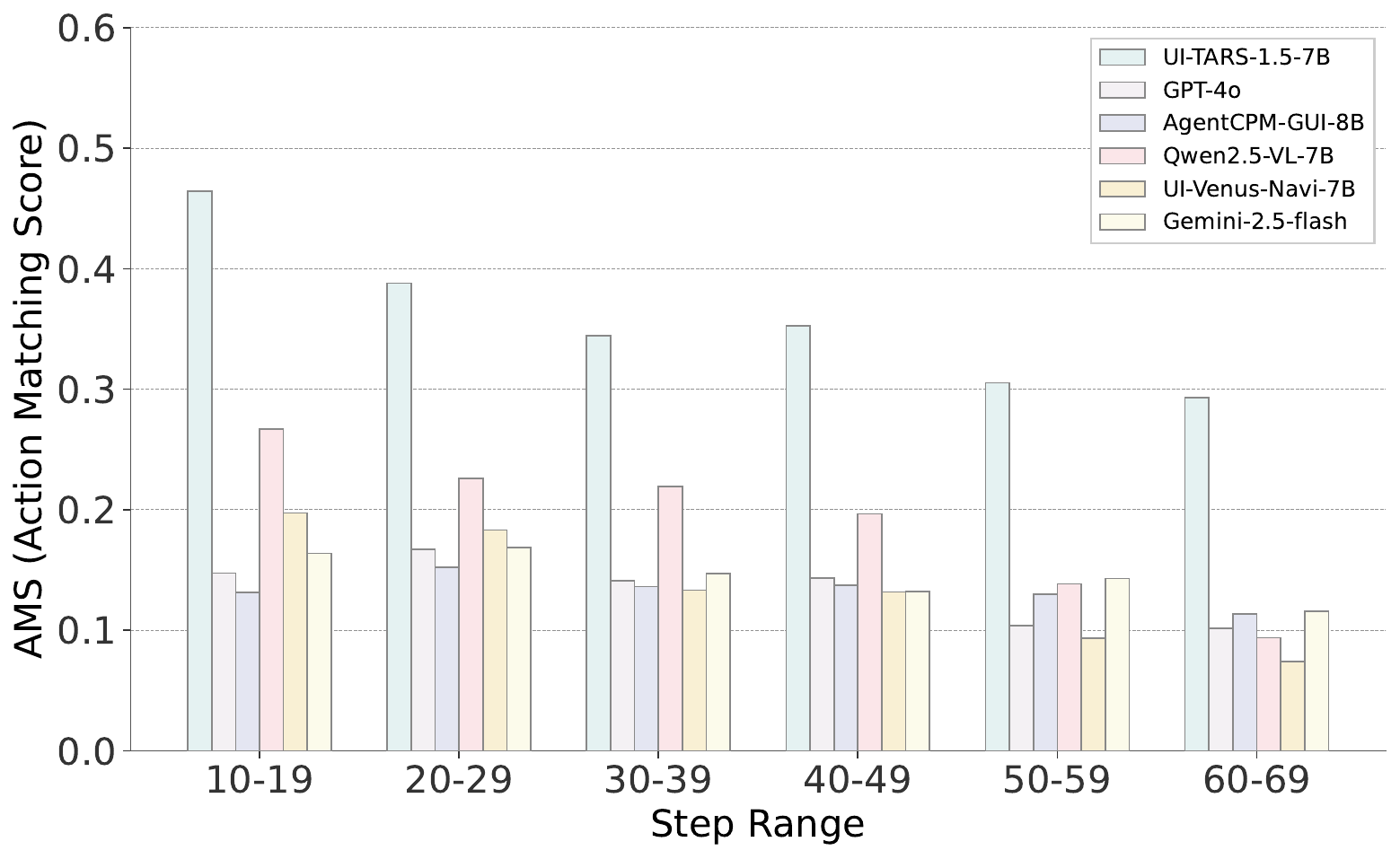}
        \caption{Raw History}
        \label{fig:placeholder1}
      \end{subfigure}
      \hfill
      \begin{subfigure}[t]{0.32\linewidth}
        \centering
        \includegraphics[width=\linewidth]{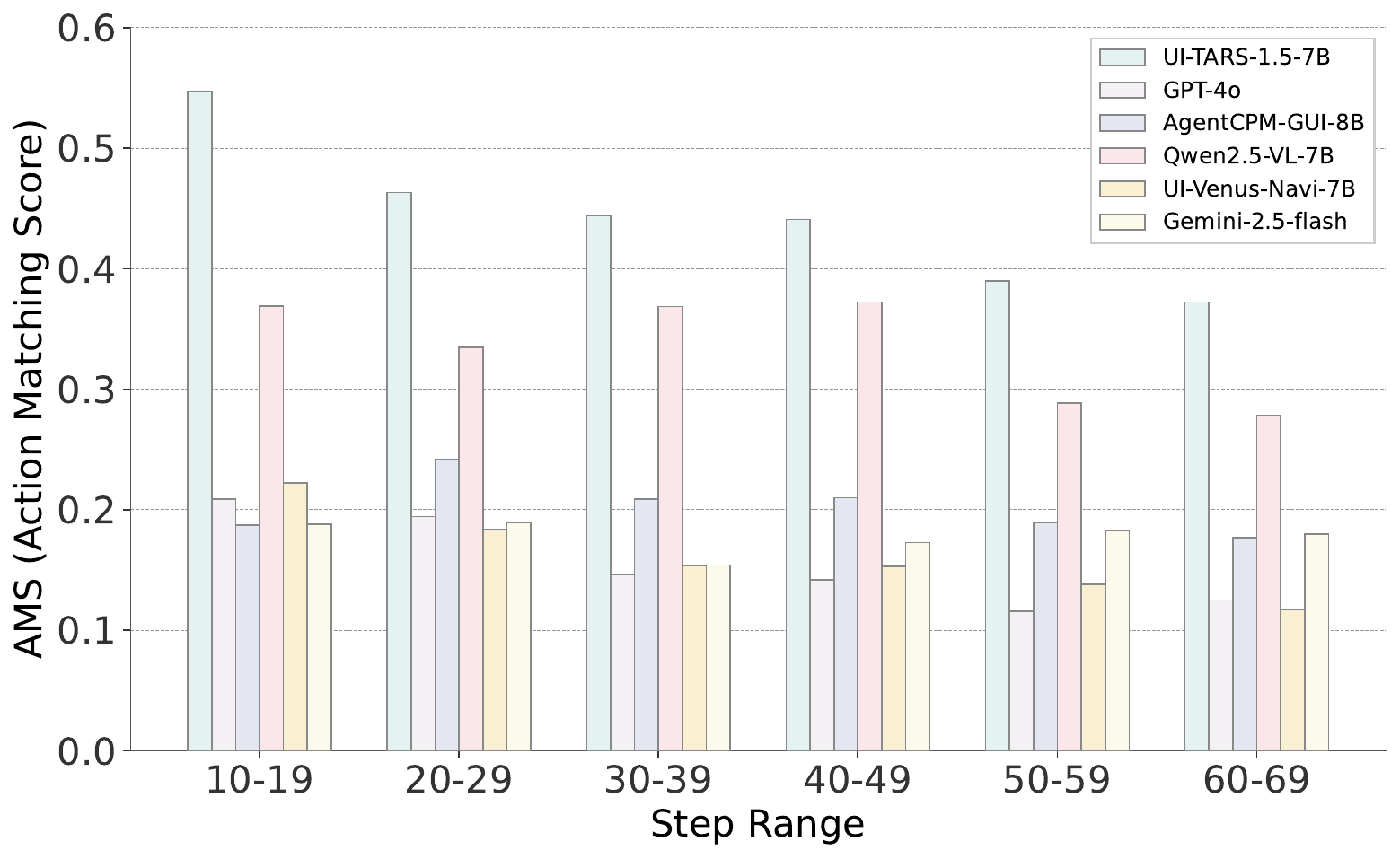}
        \caption{Coarse Summary}
        \label{fig:placeholder2}
      \end{subfigure}
      \hfill
      \begin{subfigure}[t]{0.32\linewidth}
        \centering
        \includegraphics[width=\linewidth]{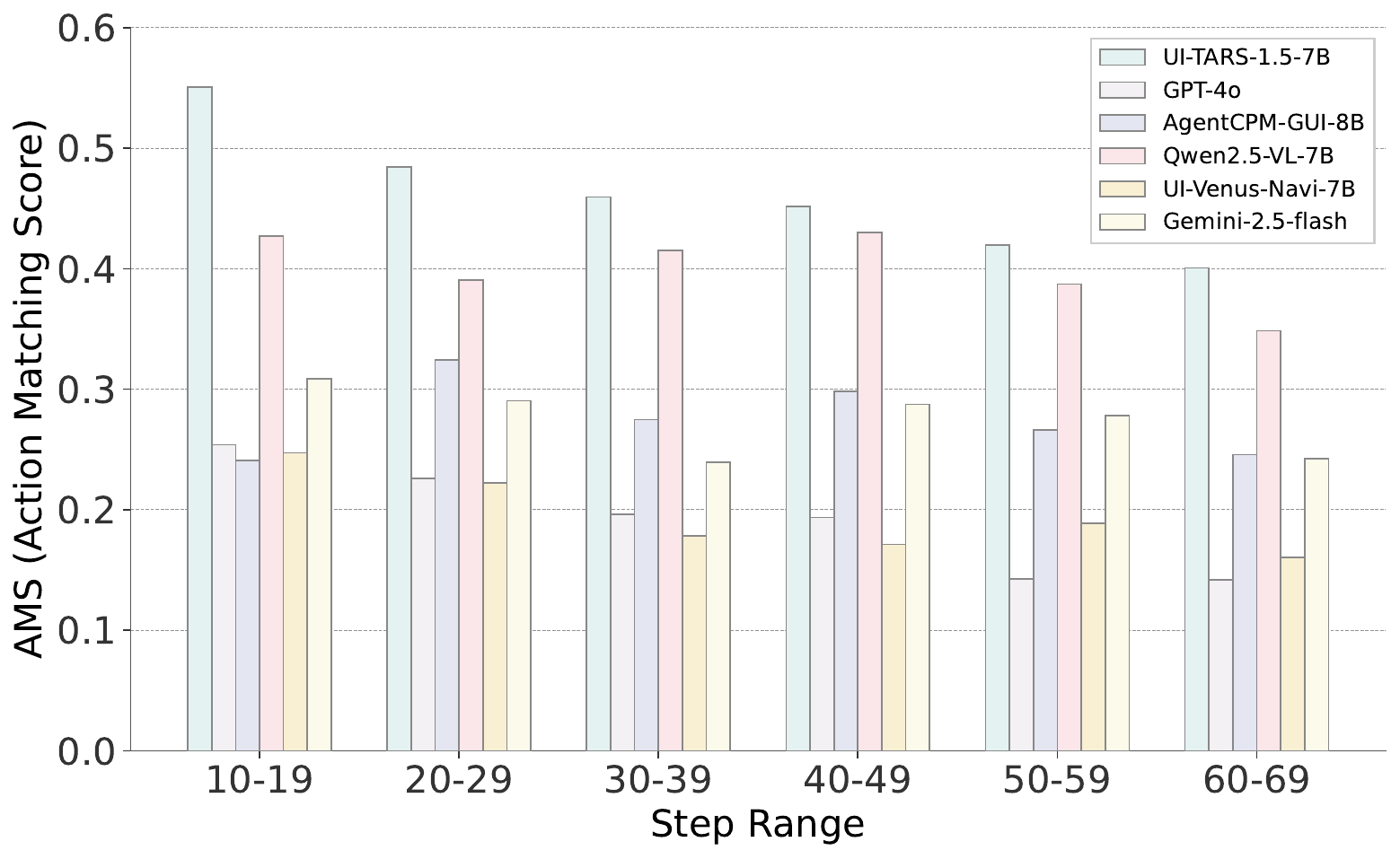}
        \caption{ASM (Ours)}
        \label{fig:placeholder3}
      \end{subfigure}
    
      \vspace{4pt}
    
    \caption{Agent performance across different interaction step ranges under three history utilization strategies: (a) Raw History, (b) Coarse Summary, and (c) \methodname{} (ASM).}
    \label{fig:5_exp_different_history_line}
\end{figure*}

Long-horizon mobile GUI tasks require agents to maintain and utilize interaction history across many steps. In \BenchName{}-Bench, tasks often span multiple applications and contain dependency chains among intermediate states (e.g., extracted information, completed subgoals, or environment changes) that determine downstream decisions.
Existing GUI agents typically rely on two forms of history representation. 
\textbf{Raw interaction traces} preserve full screenshot--action trajectories but grow rapidly with task horizon, causing irrelevant UI transitions to dilute attention over task-relevant states. 
\textbf{Compressed summaries} reduce context length but often omit fine-grained intermediate states needed for later reasoning.
Our benchmark diagnosis shows that both approaches degrade significantly as interaction horizons grow (Fig.~\ref{fig:5_exp_different_history_line}), indicating that \emph{history representation is a key bottleneck for long-horizon GUI agents}. 
These findings motivate a history representation that explicitly captures task-relevant intermediate states and their causal roles in the workflow.

\subsection{Anchored State Memory Definition}
\label{sec:6_method_definition}
% To address this limitation, we propose \textbf{Anchored State Memory (ASM)}, a structured representation that organizes interaction history into sparse but decision-critical intermediate states.

% Given an interaction trajectory

% \begin{equation}
% \tau = \{(s_t, a_t)\}_{t=0}^{T},
% \end{equation}

% where $s_t$ denotes the UI state and $a_t$ the executed action, ASM extracts a set of \textit{state anchors}

% \begin{equation}
% A = \{m_k\}_{k=1}^{K}.
% \end{equation}

% Each anchor corresponds to a task-relevant intermediate state and is represented as

% \begin{equation}
% m_k =
% \langle
% \text{type}_k,
% \text{content}_k,
% \text{evidence}_k,
% \text{links}_k
% \rangle.
% \end{equation}

% Specifically, \textit{type} indicates the functional role of the state in the workflow, \textit{content} describes the semantic information of the state, \textit{evidence} records supporting interaction steps, and \textit{links} represent causal or dependency relations with other anchors. 
% During task execution, the agent retrieves anchors conditioned on the current subgoal and context, enabling it to reference the most relevant historical states without relying on extremely long raw trajectories. 

To address this limitation, we propose \textbf{Anchored State Memory (ASM)}, a structured memory representation that organizes interaction history into sparse but decision-critical intermediate states, rather than replaying the full trajectory or compressing it into a free-form summary. ASM explicitly represents not only \emph{what} intermediate states should be remembered, but also \emph{how} they are causally related and reused across steps.

Given an interaction trajectory

\begin{equation}
\tau = \{(s_t, a_t)\}_{t=0}^{T},
\end{equation}

where $s_t$ denotes the UI state and $a_t$ the executed action, ASM extracts a set of \textit{state anchors}

\begin{equation}
A = \{m_k\}_{k=1}^{K}.
\end{equation}

Each anchor corresponds to a task-relevant intermediate state that may be required by future decisions, such as an extracted value, an identified entity, a completed subgoal, a persistent state change, or an exception that must later be handled. Each anchor is represented as

% link实际上不是锚点内部的定义，而是整体的锚点网络
\begin{equation}
m_k =
\langle
\text{type}_k,
\text{content}_k,
\text{evidence}_k,
\text{links}_k
\rangle.
\end{equation}

Specifically, $\textit{type}_k$ indicates the functional role of the state in the workflow(see Appendix~\ref{app:milestone_anchor} for the full definition of anchor types), $\textit{content}_k$ describes the semantic information carried by the state, $\textit{evidence}_k$ records the supporting UI observations or interaction steps from which the anchor is grounded, and $\textit{links}_k$ represent causal or dependency relations with other anchors.

The key distinction between ASM and conventional history representations is that ASM treats memory as a set of reusable \emph{intermediate states}, rather than as a verbatim sequence of observations and actions. This design is particularly suitable for long-horizon GUI tasks, where information obtained in one phase or app must often be reused only after many intervening steps in a later phase.
In addition to individual anchors, ASM explicitly models the dependency structure among them. A link in $\textit{links}_k$ indicates that an anchor is derived from, depends on, or should be retrieved together with another anchor. For example, in a cross-app comparison task, the anchor \textit{Cheaper\_Item} may depend on two previously extracted anchors, \textit{Price\_JD} and \textit{Price\_Taobao}; the resulting dependency relation makes this non-local reasoning step explicit and facilitates later retrieval when the agent needs to decide which item to add to the cart.

% 检索实际上是由模型自己一并检索的，在我们自己的agent上，这个检索可以独立实现
During task execution, the agent maintains a memory bank of anchors and interacts with ASM in a retrieve--reason--update manner. At each step, the agent retrieves anchors conditioned on the current UI state and user instruction, enabling it to reference the most relevant historical states without relying on extremely long raw trajectories. The retrieved anchors are then used jointly with the current UI context to predict the next action. After that, the agent analyzes the current step to determine whether new anchors should be created, whether existing anchors should be updated or invalidated, and whether new dependency links should be added to the memory bank.

Formally, at step $t$, given the current UI state $s_t$ and the current memory bank $A_{t-1}$, ASM produces a retrieved subset $\hat{A}_t \subseteq A_{t-1}$ for decision making, predicts the next action $a_t$, and then updates the memory bank:
\begin{equation}
\hat{A}_t = \mathrm{Retrieve}(s_t, A_{t-1}),
\end{equation}
\begin{equation}
a_t = \mathrm{Act}(s_t, \hat{A}_t),
\end{equation}
\begin{equation}
A_t = \mathrm{Update}(A_{t-1}, s_t, a_t).
\end{equation}

Compared with raw trajectory replay, ASM removes locally irrelevant observations and reduces distraction from long histories. Compared with free-form summaries, it preserves dependency-critical intermediate states in a grounded, structured, and updateable form. This makes ASM particularly effective for long-horizon GUI tasks that require non-local state reuse and cross-app coordination.

\setlength{\tabcolsep}{6pt}
\renewcommand{\arraystretch}{1.12}

\begin{table}[t]
\centering
\caption{History modeling ablation on \BenchName{}-Bench. Columns report AMS and TCR (higher is better, $\uparrow$). For each model, best is in \textbf{bold} and second-best is \underline{underlined} across Raw/Summary/\methodname{}.}
\label{tab:5_history_ablation_vertical}
\begin{adjustbox}{width=0.75\linewidth}
\begin{tabular}{llcccc}
\toprule
\textbf{Model} & \textbf{History} & \textbf{AMS}$_{\uparrow}$ & \textbf{TCR}$_{\uparrow}$ & \textbf{Avg Token} & \textbf{Avg Time}$_{\downarrow}$ \\
\midrule
\multicolumn{6}{c}{\textbf{Closed-Source}} \\
% \cmidrule(lr){1-6}
\midrule

\multirow{3}{*}{GPT-4o~\cite{hurst2024gpt}}
& + Raw              & 14.24 & 11.75 & 2671.1 & 7.019 \\
& + Summary          & \underline{15.49} & \underline{13.02} & \textbf{993.4} & \textbf{4.521} \\
& + \methodname{}& \textbf{19.83} & \textbf{17.72} & \underline{1265.2} & \underline{5.611} \\
\midrule

\multirow{3}{*}{GPT-5~\cite{singh2025openai}}
& + Raw              & 12.37 & 11.46 & 2676.1 & 10.465 \\
& + Summary          & \underline{18.87} & \underline{17.08} & \textbf{911.3} & \textbf{6.210} \\
& + \methodname{}& \textbf{21.63} & \textbf{23.30} & \underline{1511.9} & \underline{7.320} \\
\midrule

\multirow{3}{*}{Gemini-2.5-Flash~\cite{comanici2025gemini}}
& + Raw              & 15.62 & 14.51 & 2893.6 & 14.917 \\
& + Summary          & \underline{17.14} & \underline{16.59} & \textbf{1348.6} & \underline{9.538} \\
& + \methodname{}& \textbf{26.90} & \textbf{33.24} & \underline{1783.7} & \textbf{8.694} \\
\midrule
\multirow{3}{*}{Gemini-2.5-Pro~\cite{comanici2025gemini}}
& + Raw              & 32.71 & 41.11 & 2713.6 & 15.511 \\
& + Summary          & \underline{36.39} & \underline{40.83} & \textbf{1363.7} & \underline{10.617} \\
& + \methodname{}& \textbf{57.37} & \textbf{63.40} & \underline{1810.1} & \textbf{8.732} \\
\midrule
\multirow{3}{*}{Gemini-3-Flash}
& + Raw              & 46.14 & 55.21 & 6239.1 & \underline{8.108} \\
& + Summary          & \underline{50.05} & \underline{56.44} & \textbf{2021.3} & \textbf{5.427} \\
& + \methodname{}& \textbf{59.03} & \textbf{65.05} & \underline{2503.6} & 8.345 \\
\midrule
\multicolumn{6}{c}{\textbf{Open-Source}} \\
% \cmidrule(lr){1-6}
\midrule

\multirow{3}{*}{Qwen2.5-VL-7B~\cite{bai2025qwen2}}
& + Raw              & 20.71 & 16.04 & 13012.8 & 8.102 \\
& + Summary          & \underline{35.09} & \underline{42.39} & \textbf{5691.3} & \underline{2.922} \\
& + \methodname{}& \textbf{40.36} & \textbf{46.19} & \underline{6065.7} & \textbf{2.511} \\
\midrule

\multirow{3}{*}{UI-TARS-1.5-7B~\cite{qin2025ui}}
& + Raw              & 35.62 & 34.55 & 4762.7 & 3.109 \\
& + Summary          & \underline{44.49} & \underline{43.84} & \textbf{1343.5} & \underline{2.023} \\
& + \methodname{}& \textbf{46.07} & \textbf{46.21} & \underline{1926.7} & \textbf{1.925} \\
\midrule

\multirow{3}{*}{AgentCPM-GUI-8B~\cite{zhang2025agentcpm}}
& + Raw              & 14.06 & 10.49 & 3561.4 & 0.880 \\
& + Summary          & \underline{21.40} & \underline{22.96} & \textbf{1785.8} & \underline{0.844} \\
& + \methodname{}& \textbf{28.63} & \textbf{27.89} & \underline{1902.3} & \textbf{0.837} \\
\midrule

\multirow{3}{*}{UI-Venus-Navi-7B~\cite{gu2025ui}}
& + Raw              & 14.08 & 9.64 & 4083.8 & 3.151 \\
& + Summary          & \underline{16.07} & \underline{13.12} & \underline{1807.1} & \underline{2.578} \\
& + \methodname{}& \textbf{19.01} & \textbf{16.64} & \textbf{1186.9} & \textbf{2.343} \\
\midrule
\multirow{3}{*}{InfiGUI-R1~\cite{liu2025infigui}}
& + Raw              & 14.12 & 9.20 & 15987.3 & 5.303 \\
& + Summary          & \underline{19.84} & \underline{14.36} & \textbf{5563.0} & \textbf{2.485} \\
& + \methodname{}& \textbf{25.46} & \textbf{18.92} & \underline{6115.9} & \underline{3.424} \\

\midrule
\multicolumn{6}{c}{\textbf{Multi-Agent}} \\
% \cmidrule(lr){1-6}
\midrule

COLA~\cite{zhao2025cola} & - & 15.77 & 11.71 & 2167.1 & 14.348 \\
\midrule
Mobile-Agent-E~\cite{wang2025mobile} & - & 17.22 & 11.85 & 11099.83 & 33.960 \\
\bottomrule
\end{tabular}
\end{adjustbox}
\end{table}

\begin{figure*}[t]
    \centering
    \captionsetup[subfigure]{justification=centering}
    % Row 1
      \begin{subfigure}[t]{0.28\linewidth}
        \centering
        \includegraphics[width=\linewidth]{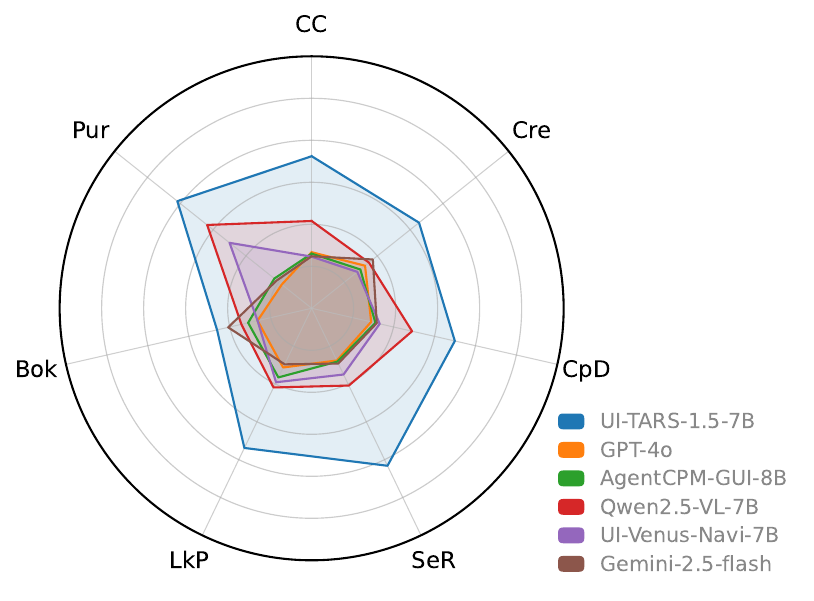}
        \caption{Raw History}
        \label{fig:placeholder4}
      \end{subfigure}
      \hspace{2mm}
      \begin{subfigure}[t]{0.28\linewidth}
        \centering
        \includegraphics[width=\linewidth]{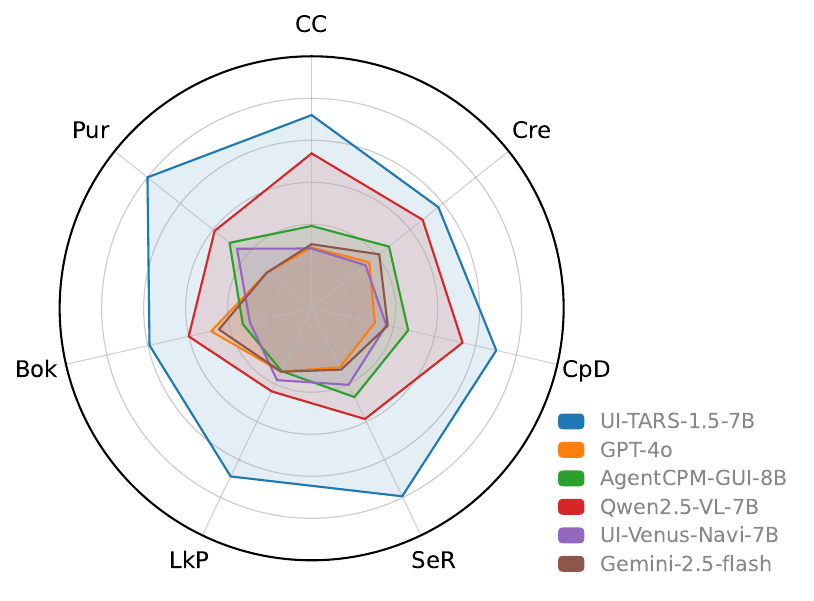}
        \caption{Coarse Summary}
        \label{fig:placeholder5}
      \end{subfigure}
      \hspace{2mm}
      \begin{subfigure}[t]{0.28\linewidth}
        \centering
        \includegraphics[width=\linewidth]{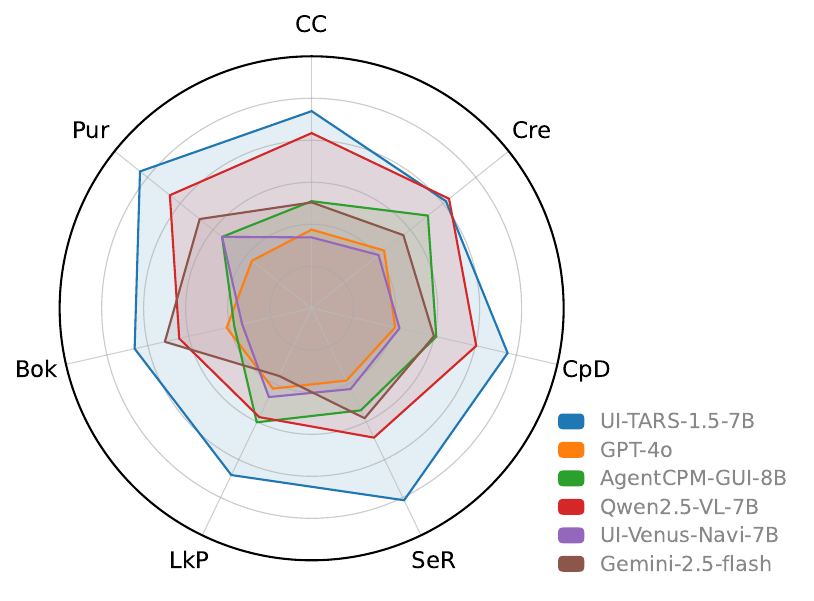}
        \caption{ASM(ours)}
        \label{fig:placeholder6}
      \end{subfigure}
    \caption{Model performance across different task categories under three history utilization strategies.}
    \label{fig:5_exp_different_history_radar}
    % \vspace{-2em}
\end{figure*}

\subsection{History Utilization Ablation: Does ASM Help?}
\label{sec:6_setup_history}
\paragraph{Setup (History ablation).}
To study how interaction history representation affects long-horizon decision making, 
we evaluate three history utilization mechanisms that differ only in how historical information is represented:

(i) \textbf{Raw History}, where the agent receives the sequence of past screenshot–action pairs;

(ii) \textbf{Coarse Summary}, where the agent is provided with a compressed textual summary of prior interactions;

(iii) \textbf{\methodname{}}, where task-relevant intermediate states and their causal relations are represented as structured milestone anchors.

Importantly, we do not provide manually crafted summaries or anchors. 
Instead, both summaries and anchors are generated automatically by each agent from the raw interaction history using the same set of rules. 
This design avoids biases introduced by artificially controlling the information strength and ensures that differences across mechanisms primarily arise from how interaction history is represented and utilized.

% \paragraph{Main Results.}
% \paragraph{Overall Performance.}

\noindent \textit{Overall Performance.} Table~\ref{tab:5_history_ablation_vertical} reports the results across multiple GUI agents. Across both closed-source and open-source models, ASM consistently achieves the best performance in terms of AMS and TCR.
These improvements demonstrate that explicitly modeling intermediate states substantially enhances agents' ability to perform long-horizon reasoning.

% \paragraph{Robustness to Long Horizons.}

% \noindent \textit{Robustness to Long Horizons.} To analyze how history representations scale with task length, we evaluate AMS across different step ranges. As shown in Fig.~\ref{fig:5_exp_different_history_line}, both raw traces and summaries exhibit noticeable performance degradation as the step range increases. In contrast, ASM maintains significantly higher AMS across long horizons, indicating stronger robustness to history accumulation.

% % \paragraph{Task Category Analysis.}

% \noindent \textit{Task Category Analysis.} We further analyze model performance across different task categories (Fig.~\ref{fig:5_exp_different_history_radar}). ASM consistently improves results across categories, particularly in tasks requiring cross-step reasoning or cross-app coordination.

\noindent \textit{Robustness and Task-Level Analysis.} We analyze how history representations scale with task length and generalize across task categories. As shown in Fig.~\ref{fig:5_exp_different_history_line}, both raw traces and summaries exhibit noticeable performance degradation as the step range increases, while ASM maintains significantly higher AMS across long horizons, indicating stronger robustness to history accumulation. Across task categories (Fig.~\ref{fig:5_exp_different_history_radar}), ASM consistently improves performance, with particularly notable gains in tasks requiring cross-step reasoning or cross-app coordination.

% \paragraph{Efficiency Analysis.}

\noindent \textit{Efficiency Analysis.} 
% We further examine the execution cost of different history modeling strategies in Table~\ref{tab:5_history_ablation_vertical}. While ASM consistently outperforms summary-based and raw-history baselines in AMS and TCR, its token usage and inference time remain in the same order of magnitude as summary-based history modeling, and are generally much lower than those of raw trajectory replay. This indicates that the gains of ASM do not come from substantially increased inference budget; instead, they stem from a more effective organization of historical information.
Table~\ref{tab:5_history_ablation_vertical} also reports execution cost. ASM consistently improves AMS and TCR over summary-based and raw-history baselines, while keeping token usage and inference time comparable to summary-based history and far below raw trajectory replay, suggesting a better efficiency--effectiveness trade-off.

% Long-chain Robustness under different history

% \paragraph{Failure Mode Analysis.}

\subsection{Failure Modes: What goes wrong in long-horizon history?}
\label{sec:6_failure_modes}
% 附录里要加
\begin{figure}[t]
    \centering
    \includegraphics[width=\linewidth]{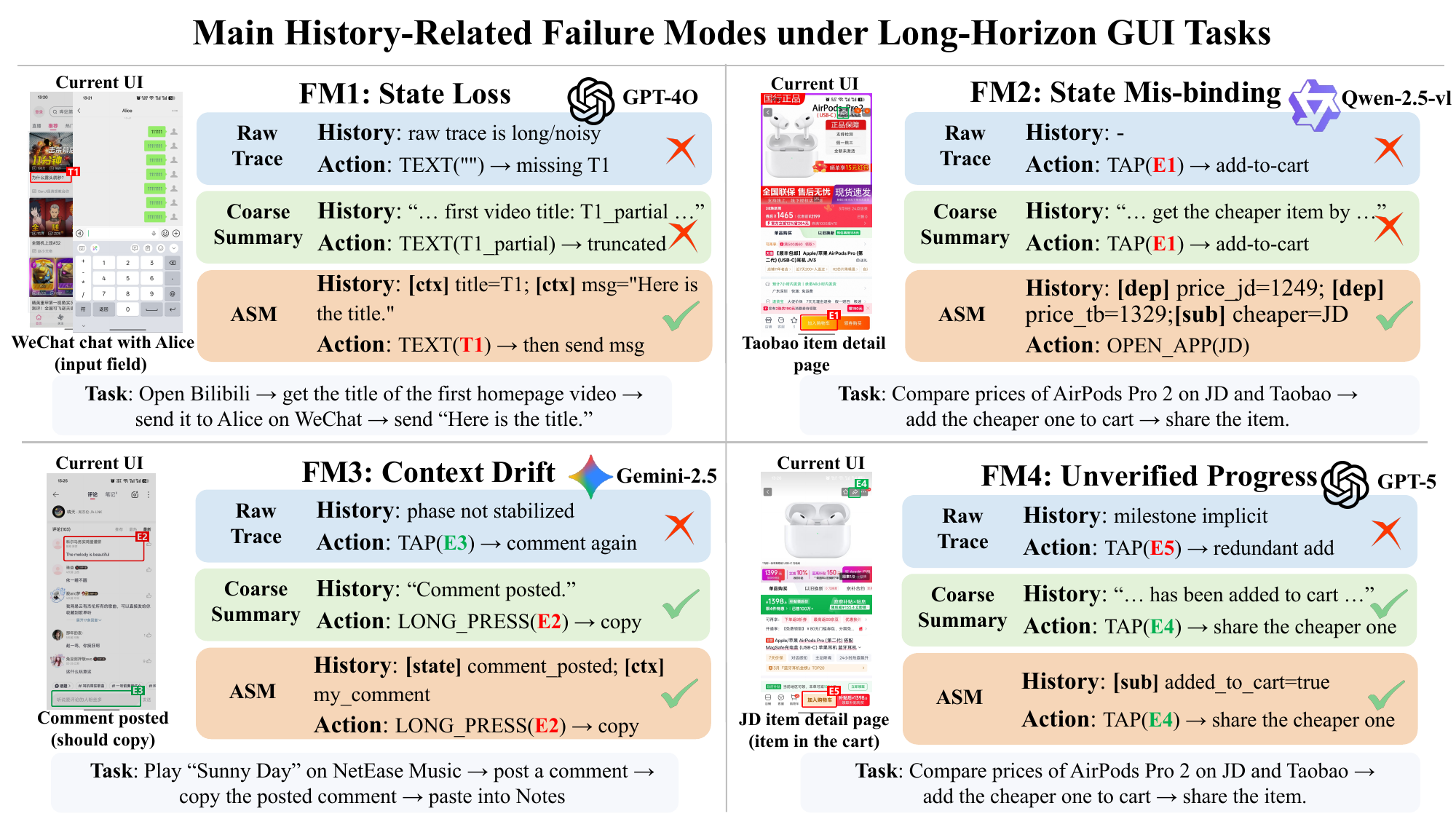}
    \caption{History-related failure modes in long-horizon GUI tasks. 
Raw traces and coarse summaries often lose task-relevant states, leading to errors such as state loss, mis-binding, context drift, and unverified progress, while ASM preserves structured intermediate states for correct decisions.}
    \label{fig:6_case}
    % \vspace{-2em}
\end{figure}
Beyond aggregate performance metrics, we analyze common failure modes observed in long-horizon interactions. These failure modes help explain why raw traces and coarse summaries degrade more rapidly than ASM as horizons grow. 
As illustrated in Fig.~\ref{fig:6_case}, across models and tasks, failures frequently arise from difficulties in maintaining and reusing task-critical intermediate states.

We identify several recurring patterns:

(i) \textbf{State Loss}, where task-relevant information extracted earlier in the trajectory becomes unavailable when needed later;

(ii) \textbf{State Mis-binding}, where intermediate states are associated with incorrect UI elements or objects;

(iii) \textbf{Context Drift}, where agents misinterpret the current interaction context or application state;

(iv) \textbf{Unverified Progress}, where agents proceed without confirming that a subgoal has been successfully completed;

(v) \textbf{Interruption Handling Failure}, where agents fail to recover from pop-ups, permission dialogs, or unexpected UI interruptions.

These failure modes frequently occur in long interaction trajectories and highlight the importance of explicitly modeling intermediate states and their dependencies. \methodshortname  mitigates these issues by organizing task-relevant intermediate states into structured anchors with causal relations, enabling more stable decision making across extended interaction sequences.
\section{Limitations}
While AndroTMem isolates and diagnoses within-task interaction memory under long-horizon Android trajectories, it does not yet fully capture \emph{cross-session} long-term tasks where critical state must persist across interruptions, days-long gaps, or multi-episode workflows. In addition, our benchmark is built on curated trajectories with fixed task goals and relatively stable app states; future progress will benefit from \emph{more dynamic} benchmarks that expose UI drift, content updates, and stochastic outcomes, as well as \emph{environment-oriented} evaluations that more tightly couple agent behavior with evolving external conditions and feedbacks.
\section{Conclusion}
\label{sec:6_conclusion}

In this paper, we introduce AndroTMem-Bench, a benchmark for studying long-horizon mobile GUI tasks with cross-app workflows and complex dependency structures. Our diagnosis shows that existing GUI agents degrade significantly as task horizons increase, revealing history modeling as a key bottleneck.
To address this issue, we propose \textbf{Anchored State Memory (ASM)}, a structured history representation that organizes interaction trajectories into sparse intermediate-state anchors and their causal relations.
Experiments across diverse GUI agents show that ASM consistently improves both action accuracy and task completion, while enhancing robustness in long-horizon scenarios. Our analysis further identifies common history-related failure modes and demonstrates how structured intermediate-state memory mitigates them.
We hope AndroTMem and ASM facilitate future research on memory mechanisms for reliable long-horizon GUI agents.

\clearpage

\newpage
\bibliography{references}

\clearpage
% \clearpage
% \setcounter{page}{1}
% \title{AndroTMem: From Interaction Trajectories to Anchored Memory in Long-Horizon GUI Agents Supplementary Material} 
% \maketitlesupplementary
% \appendix

\newpage
\appendix
\begin{center}
\Large
\textbf{Appendix}
\end{center}
\vspace{-3mm}
\addcontentsline{toc}{section}{Appendix} 

% You can have as much text here as you want. The main body must be at most $8$
% pages long. For the final version, one more page can be added. If you want, you
% can use an appendix like this one.

% The $\mathtt{\backslash onecolumn}$ command above can be kept in place if you
% prefer a one-column appendix, or can be removed if you prefer a two-column
% appendix.  Apart from this possible change, the style (font size, spacing,
% margins, page numbering, etc.) should be kept the same as the main body.
\section{Details of \BenchName{}-Bench}
\label{app:androtrace}
\subsection{App Categories}
\label{app:androtrace_app_categories}
To cover a broad spectrum of mobile interaction scenarios, the applications in our study are organized into 16 task-oriented functional domains. These categories represent the most frequent and essential user activities within the Android ecosystem, encompassing a broad spectrum of UI structures and interaction logic—from simple utility functions to complex social and productivity workflows. The detailed classification, representative applications, and their corresponding functional descriptions are systematically documented in Table \ref{tab:app_categories_1}.

\begin{table*}[h]
\centering
\caption{Functional app grouping based on predefined task-oriented categories (Part I).}
\label{tab:app_categories_1}
\scriptsize
\setlength{\tabcolsep}{4pt}
\renewcommand{\arraystretch}{1.05}
\begin{tabular}{@{}p{0.17\textwidth} p{0.29\textwidth} p{0.48\textwidth}@{}}
\toprule
Category & Applications & Description \\
\midrule
Shopping & JD, Taobao, Pinduoduo, Suning & Product browsing, price comparison, order placement, and online payment for consumer goods. \\
Travel & Ctrip, Didi Chuxing, Fliggy & Trip planning, transportation booking, ride-hailing services, and real-time travel management. \\
Map & Amap, Baidu Maps, Tencent Maps & Location search, route planning, navigation, and real-time traffic information. \\
Mail & Gmail, Outlook & Email communication, message management, information exchange, and productivity-oriented correspondence handling. \\
Cloud Storage & Baidu Netdisk, Google Drive & Cloud-based file storage, synchronization, backup, and cross-device file sharing. \\
Meeting & DingTalk, Tencent Meeting, WeCom, Zoom & Online meetings, video conferencing, real-time communication, and collaborative work coordination. \\
Note-taking \& Document Tools & Feishu (Lark), WPS Office, DingTalk, Notes, Youdao Notes & Document editing, note-taking, information organization, and collaborative content creation. \\
Social & WeChat, QQ, Weibo, Xiaohongshu & Social networking, instant messaging, content sharing, and online community interaction. \\
Category & Applications & Description \\

Entertainment & Bilibili, Douyin, Tencent Video, iQIYI, YouTube & Consumption of short-form and long-form video content, live streaming, and entertainment media browsing. \\
News \& Information & Toutiao, NetEase News, Sina News & News aggregation, personalized information feeds, and real-time information consumption. \\
System Tool & File Manager, Gallery, Settings, Browser, Recorder, Calculator, Clock, Calendar, Chrome & Device management, system configuration, file organization, and basic utility functions. \\
Knowledge & Zhihu & Knowledge sharing, question answering, expert content consumption, and community-based discussions. \\
Music App & NetEase Cloud Music, QQ Music, Kugou Music, Kuwo Music & Music streaming, playlist management, audio search, and personalized music recommendation. \\
Takeout & Meituan, Ele.me & Food delivery services, restaurant browsing, order placement, and logistics tracking. \\
Payment & Alipay & Mobile payment, financial transactions, digital wallet management, and online-offline payment integration. \\
Health & Keep & Fitness training, exercise tracking, health monitoring, and personalized workout guidance. \\
\bottomrule
\end{tabular}
\end{table*}

\subsection{Task Types}
\label{app:androtrace_task_types}
In \BenchName{}-Bench, long-chain tasks are categorized according to the user's primary intent rather than application domains. This approach acknowledges that even within the same application, user goals vary significantly and dictate different interaction patterns. In \BenchName{}-Bench, each task is mapped to a single primary intent that represents the core objective of the sequence. Table~\ref{tab:task_types_daily} details the classification of simulated daily operations, providing definitions and representative examples for each category.

% 引用~\ref{tab:actions}
% To facilitate precise GUI navigation and trajectory annotation, we define a comprehensive action space $\mathcal{A}$ that encapsulates the fundamental interactions between the agent and the Android environment. These actions are dispatched via the Android Debug Bridge (ADB), enabling the system to simulate a unified set of interactions spanning different abstraction levels, ranging from low-level touch gestures (e.g., tapping or swiping at specific screen coordinates) to high-level system commands (e.g., app launching or navigation controls). Each action $a \in \mathcal{A}$ is defined by its operation type and associated parameters (e.g., screen coordinates or text strings). The detailed taxonomy of the action space in AndroTrace, including their respective arguments and functional descriptions, is formalized in Table~\input{tables/9_action_space}.
A parameterized action space $\mathcal{A}$ is defined to support precise GUI interaction and trajectory annotation in \BenchName{}-Bench. The action space provides a unified interface that covers both low-level touch-based interactions and high-level system commands on Android devices.
Each action $a \in \mathcal{A}$ is specified by an operation type and associated arguments. Table~\ref{tab:appendix_action_space} summarizes the action space used in \BenchName{}-Bench, detailing the supported action types together with their required arguments and functional descriptions.%如果数学符号表达mathcal{A}这种是不需要的，可以直接删了
\noindent
We use a unified coordinate system and consistent argument schema so that the same action definitions apply across devices and screen resolutions.
All actions are logged with their parameters to support reproducible replay and fine-grained analysis.

\begin{table*}[ht]
\centering
\caption{Classification of tasks in \BenchName{} based on primary user intent.}
\label{tab:task_types_daily}
\small
\setlength{\tabcolsep}{4pt}
\renewcommand{\arraystretch}{1.3}
\begin{tabular}{@{}p{0.18\textwidth} p{0.24\textwidth} p{0.53\textwidth}@{}}
\toprule
\textbf{Task Type} & \textbf{Primary Intent} & \textbf{Example Scenarios} \\ \midrule
Lookup & Information retrieval/verification & Checking shop hours, searching for news, or finding answers to specific questions. \\

Compare \& Decide & Decision making via comparison & Comparing menus of three restaurants or checking product parameters across different platforms. \\

Purchase / Order & Transaction completion & Placing food delivery orders, clearing a shopping cart, or purchasing memberships. \\

Booking & Resource/Time reservation & Booking hotel rooms, reserving train tickets, or scheduling a doctor's appointment. \\

Communicate & Coordination with others & Confirming meeting details with a contact or initiating a group discussion. \\

Share / Recommend & Content distribution & Sending a store link to a friend or sharing news and music playlists to social media. \\

Create Content & Generative output & Writing a note, editing a document, or posting a review/comment on a platform. \\

Configure & System/App adjustment & Granting location permissions, enabling notifications, or modifying system settings. \\

Payment & Financial processing & Paying for an existing order, applying for a refund, or processing utility bills. \\

Navigation & Trip execution & Planning a route between locations or hailing a ride and sharing the trip status. \\ \bottomrule
\end{tabular}
\end{table*}

\begin{table*}[t]
\centering
\caption{The argument and functionality of different actions in \BenchName{}. 'pos1' and 'pos2' denote the position $(x, y)$.}
\label{tab:appendix_action_space}
\small
\setlength{\tabcolsep}{6pt}
\renewcommand{\arraystretch}{1.2}

\begin{tabular}{@{}p{0.20\textwidth} p{0.12\textwidth} p{0.60\textwidth}@{}}
\toprule
Action & Argument & Functionality \\
\midrule
\texttt{TAP} & [pos1] &
click the on-screen position \\

\texttt{INPUT\_TEXT} & [text] &
input the text into the currently focused text field using the system keyboard \\

\texttt{LONG\_PRESS} & [pos1] &
press the screen for a long time to copy texts or download images \\

\texttt{SWIPE} & [pos1] &
swipe the screen in a given direction (up, down, left, or right) to scroll the page \\

\texttt{SWIPE\_TWO\_POINT} & [pos1,pos2] &
swipe from position pos1 to position pos2, enabling precise scrolling or drag-and-drop operations \\

\texttt{WAIT} & -- &
wait for a specified duration or until the UI state changes \\

\texttt{FINISH} & -- &
the sign that the instruction has been completed \\

\texttt{OPEN\_APP} & -- &
launch the specified application \\

\texttt{CAPTURE\_SCREEN} & -- &
capture the current screen for visual perception or state analysis \\

\texttt{HOME} & -- &
go to the home screen \\

\texttt{BACK} & -- &
go to the previous screen \\
\bottomrule
\end{tabular}
\end{table*}

\subsection{Action Space}
\label{app:androtrace_app_set}

\subsection{Processing of Private Information}
\label{app:androtrace_privacy}
% 标注过程中，通过建立临时账号/修改信息的方式，去除标注过程中任何可能标识身份的信息（地址、账户名、手机号、联系人、付款信息等）
All personally identifiable information (PII) is strictly anonymized during the annotation process to ensure privacy protection. This was achieved by utilizing temporary accounts and systematically modifying or replacing sensitive data fields. Specifically, any information that could potentially identify an individual—including residential addresses, account usernames, phone numbers, contact lists, and financial payment details—was entirely removed or substituted with generic placeholders. Consequently, the traces within \BenchName{}-Bench contain no traceable real-world identity markers.
\subsection{Task Template Examples}
\label{app:androtrace_template}
% 任务指令模板示例
% #表示从固定的可选项中填充，#group是预设的APP功能组划分（有一张表,appendix_a_app_classification），#contract是从预设的联系人列表中选择
% Compare the prices of the best-selling {product that can be uniquely identified} on {#group:shopping_app}, {#group:shopping_app}. Add the product with the highest price to your shopping cart and share it with your contact {#contract} on {#group_social}.
% 填充示例
% Compare the prices of the best-selling Apple AirPods Pro 2 on JD, Taobao. Add the product with the highest price to your shopping cart and share it with your contact xiaoli on wechat.
% 
To ensure structural consistency while maintaining data diversity, task instructions in \BenchName{}-Bench are generated using a templated approach. Placeholders prefixed with \texttt{\#} indicate slots filled from predefined sets: \texttt{\#group} refers to specific application functional categories (as detailed in Table~\ref{tab:app_categories_1}), and \texttt{\#contract} represents a selection from a curated list of virtual contacts. An example of a complex, cross-app task template and its instantiated version is shown below:
% wujie revised
% \begin{quote}
%     \textbf{Template:} Compare the prices of the best-selling \textit{\{product\}} on \texttt{\#group:shopping\_app} and \texttt{\#group:shopping\_app}. Add the product with the highest price to your shopping cart and share it with your contact \texttt{\#contract} on \texttt{\#group:social}.

%     \textbf{Example Instance:} Compare the prices of the best-selling Apple AirPods Pro 2 on \textit{JD} and \textit{Taobao}. Add the product with the highest price to your shopping cart and share it with your contact \textit{Xiaoli} on \textit{WeChat}.
% \end{quote}
\begin{quote}\raggedright
\textbf{Template:} Compare the prices of the best-selling \textit{\{product\}} on \texttt{\#group:}\texttt{shopping\_app} and \texttt{\#group:}\texttt{shopping\_app}. Add the product with the highest price to your shopping cart and share it with your contact \texttt{\#contract} on \texttt{\#group:}\texttt{social}.

\textbf{Example Instance:} Compare the prices of the best-selling Apple AirPods Pro 2 on \textit{JD} and \textit{Taobao}. Add the product with the highest price to your shopping cart and share it with your contact \textit{Alice} on \textit{WeChat}.
\end{quote}

\subsection{Annotation Platform}
\label{app:anno_platform}
We develop a semi-automatic annotation platform based on the Android Debug Bridge (ADB) to support efficient collection and fine-grained labeling of long-horizon GUI trajectories. The platform can connect to either real Android devices or Android emulators, capture the current screen in real time, and allow annotators to label interaction actions such as \act{tap}, \act{long\_press}, \act{swipe} based on the screenshot. The annotated action commands are synchronously executed on the connected device, and the resulting device UI state is immediately reflected back to the annotator to support subsequent interactions and annotations, forming a closed-loop human-in-the-loop workflow.

In addition to recording the screenshot and corresponding action information at each step, the platform also synchronously captures the XML representation of the current UI screen (derived from the Android accessibility tree), which contains the page layout structure and UI element attributes. More importantly, the platform supports configurable auxiliary annotation fields, covering a wide range of data types, including free-form text, boolean values, JSON objects, object arrays, single-choice groups, and multiple-choice groups. 
% Leveraging this functionality, we additionally annotate, for each step, step-level reasoning traces (Reasoning, text), step-level summaries (Summary, text), and milestone anchors (Milestone Anchors, object arrays).
Leveraging this functionality, we annotate for each step a reasoning trace, a short summary, and \emph{\StateAnchors{}} (object arrays) that capture task-relevant intermediate states and their causal roles.

\section{\methodname}
\label{app:milestone_anchor}
\subsection{State Anchor Categories}
\label{app:ma_categories}
% wujie revised
Anchored State Memory (ASM) organizes interaction history into sparse intermediate-state anchors together with their causal dependency links. 
Based on their semantic roles and how they affect subsequent decisions, we categorize state anchors into six types. We summarize the taxonomy in Table~\ref{tab:milestones} and further elaborate below.

\begin{table}[H]%[t]换成了[H]来固定,并且去掉*不然的话不支持
\centering
\caption{Causal state anchors representing key task progress and execution states in GUI agent interactions.}
\label{tab:milestones}
\small
\setlength{\tabcolsep}{6pt}
\renewcommand{\arraystretch}{1.2}

\begin{tabular}{
@{}
>{\raggedright\arraybackslash}m{0.20\textwidth}
>{\raggedright\arraybackslash}m{0.20\textwidth}
>{\raggedright\arraybackslash}m{0.48\textwidth}
@{}
}
\toprule
Type Name & Recommended Values & Description \\
\midrule
Key Subgoal Achievement & \texttt{SUBGOAL} &
Completion of intermediate tasks, such as successful login, retrieval of search results, or adding an item to the shopping cart \\

State Transition and Mode Switching & \texttt{STATE\_CHANGE} &
Changes in the environment or application mode, such as entering the payment interface or opening a permission request page \\

Causally Dependent Steps & \texttt{DEPENDENCY} &
Steps whose execution depends on prior data or states, such as copying text or filling in a form \\

Exception Handling & \texttt{EXCEPTION} &
Actions taken to address interruptions or errors, such as closing pop-up advertisements or retrying a network request \\

Global Constraint Information & \texttt{CONTEXT\_INFO} &
Essential contextual information that constrains task execution, such as selecting dates or setting filter conditions \\

Task Completion Indicator & \texttt{FINISH} &
The sign that the instruction has been completed \\
\bottomrule
\end{tabular}
\end{table}

\paragraph{Subgoal completion (\act{subgoal}).}
 These anchors mark the successful completion of intermediate objectives (i.e., meaningful progress toward the overall task), such as ``successfully logged in'', ``search results displayed'', or ``item added to cart''. Subgoal anchors indicate that the agent has reached a verifiable intermediate milestone that may be required by later steps.

\paragraph{State transition and mode switching (\act{state\_change}).}
These anchors capture significant changes in application state or operating mode that alter the execution context for subsequent steps. Examples include entering a new module, switching accounts, granting permissions, or enabling a critical setting. Such transitions often change what actions are valid or what information is visible.

\paragraph{Causal dependency steps (\act{dependency}).}
These anchors denote steps whose outcomes are explicitly required by later actions. This is especially common in cross-app workflows, e.g., copying content in App~A that will be pasted in App~B, or setting a filter that directly affects subsequent results. Dependency anchors make these prerequisites explicit and support downstream validation.

\paragraph{Exception handling (\act{exception}).}
These anchors record the handling of disruptive events that must be resolved to proceed, such as closing pop-ups, dismissing ads, handling error dialogs, or retrying after network failures. Exception anchors help agents recognize abnormal conditions and apply recovery strategies instead of derailing the main workflow.

\paragraph{Global contextual information (\act{context\_info}).}
These anchors store task-critical context that may not correspond to a single atomic action but is essential for subsequent reasoning, such as key user-provided parameters (dates, locations, names) or important system feedback (warnings, error messages). They function as persistent constraints or references throughout the task.

\paragraph{Task completion (\act{finish}).}
These anchors indicate that the overall task has been successfully completed, e.g., ``order submitted'' or ``message sent''. Finish anchors define terminal success conditions and provide a clear endpoint for evaluation.
\subsection{Generation Protocol for Summary and ASM}
\label{app:exp_generation_protocol}

To ensure fairness and reproducibility in the comparison between \textit{Raw History}, \textit{Summary}, and \textit{ASM}, all historical representations are generated automatically from the same interaction trajectory using a unified rule-based protocol. Importantly, the generation process is \textbf{model-agnostic} and does not involve model-specific tuning.

\paragraph{Unified Prompting Protocol.}
Both summaries and anchors are generated using a fixed prompt template that specifies the available action space, the normalized coordinate system, the output schema, and the generation guidelines. The prompt structure is identical across all evaluated agents. The complete prompt templates used in our experiments are provided in Table~\ref{table:history_prompt} and Table~\ref{table:milestone_prompt}. These templates define the allowed actions, field requirements for action parameters, structured output formats, and semantic guidelines for summarization and anchor extraction. As a result, all models operate under the same instruction format when generating summaries or anchors.

% \paragraph{Output Schema and Format Constraints.}
% To ensure consistent outputs across models, the generated summaries and anchors follow a fixed JSON schema. For summaries, the model returns a JSON object containing the predicted action and a concise textual abstraction of the current task state (the \texttt{summary\_en} field). For anchors, the output additionally includes the semantic anchor description (\texttt{content\_en}), an explanation of its role in the workflow (\texttt{description\_en}), and an optional causal link to a previous anchor. 

% All outputs must conform to the predefined schema. If the returned result does not match the required JSON structure, the generation step is automatically retried using the same prompt until a valid output is produced. This mechanism reduces formatting variability across models and ensures consistent structured outputs.

\paragraph{Trigger Conditions for Anchor Generation.}
Anchors are generated incrementally during task execution. During evaluation, anchors are generated online by the agent from the observed interaction history rather than taken from the annotated ground truth. At each interaction step, the model receives the current UI observation, previously generated anchors, and the interaction history. Based on the generation guidelines defined in the prompt, the model decides whether a new anchor should be produced. Anchors are generated only when the interaction produces a semantically meaningful task state, such as subgoal completion, state transitions, dependency establishment, exception handling, or contextual information updates. This design ensures that anchors represent task-relevant milestones rather than trivial UI transitions.
\paragraph{Causal Link Generation.}
In addition to individual anchors, ASM explicitly models the causal dependencies between anchors. Each anchor may optionally include a causal link that connects it to a previously generated anchor, forming a lightweight dependency structure over the interaction trajectory. These links capture task-relevant relations such as prerequisite conditions, enabling states, or results of previous actions. The goal of causal links is not to construct a full execution graph, but to expose task-critical dependencies that guide memory retrieval during long-horizon interaction.
During generation, the model is instructed to produce causal links only when a semantically meaningful dependency exists between two anchors. This design avoids constructing dense or trivial connections and instead focuses on decision-critical dependencies.
By organizing anchors together with their causal links, ASM forms a structured interaction memory rather than a flat list of events. This allows the agent to retrieve only causally relevant anchors instead of processing the entire interaction history. During action prediction, the agent can retrieve anchors that are causally relevant to the current subgoal or context, enabling more targeted use of historical information.
\paragraph{Robustness to Model Output Style.}
Different models may exhibit varying abilities in structured text generation. To mitigate this effect, we apply three normalization steps: (1) unified prompting across all models, (2) strict schema enforcement for generated outputs, and (3) an automatic retry mechanism for invalid outputs. These steps ensure that the anchor generation process primarily reflects the model's interpretation of the interaction state rather than stylistic differences in output formatting.

\paragraph{Why Improvements Are Not Due to Anchor Writing Ability.}
Although anchors are generated by the evaluated models themselves, the improvements observed with ASM cannot be attributed solely to differences in anchor writing ability. First, all history representations (raw history, summary, and anchors) originate from the same interaction trajectory. Second, the same model generates both summaries and anchors using the same prompting framework. Third, consistent improvements are observed across both open-source and closed-source models, suggesting that the gains primarily arise from the structured memory representation introduced by ASM rather than model-specific generation behavior.
\subsection{Fairness of History Representation}
\label{app:exp_fairness_history_representation}
A potential concern when comparing different history representations is whether one representation implicitly provides stronger supervision or more information to the agent. To ensure a fair comparison between \textit{Raw History}, \textit{Summary}, and \textit{ASM}, we enforce the following principles.

\paragraph{Same Information Source.}
All history representations are derived from the \textbf{same interaction trajectory}. Specifically, the underlying information consists of the sequence of past UI observations and executed actions. Neither summaries nor anchors introduce additional external annotations or ground-truth signals. Instead, they are automatically generated by the agent itself based solely on the observed interaction history.

\paragraph{No Additional Evidence Signals.}
ASM does not introduce additional perceptual evidence beyond what is already available in the raw interaction history. The evidence fields in ASM simply reference observations present in the trajectory (e.g., screenshots, UI elements, or extracted values) rather than providing new information. Consequently, ASM reorganizes existing observations into a structured representation instead of injecting extra supervision. Both summaries and ASM serve as compact abstractions of the same trajectory and are substantially shorter than the raw interaction history.

\paragraph{Consistent Observation Inputs.}
At each interaction step, the agent receives the same primary observation inputs regardless of the history representation strategy: the current UI state and the user instruction. The only difference lies in how past interactions are represented in the context (raw trajectory, textual summary, or structured anchors).

\paragraph{Representation vs. Information Strength.}
The goal of ASM is not to increase the amount of information available to the agent, but to reorganize previously observed information into semantically meaningful intermediate states and their causal dependencies. Therefore, the comparison isolates the effect of \textit{history representation structure} rather than differences in information strength.
% \paragraph{Role of ASM in the Benchmark.}
% It is important to note that this work primarily introduces a diagnostic benchmark for studying interaction memory in long-horizon GUI agents. ASM is designed as a principled testbed method that operationalizes the diagnosed memory bottleneck and enables controlled evaluation of different history representations. 
% ASM should therefore be viewed as a structured baseline that exposes how explicitly representing intermediate states and their causal dependencies can improve long-horizon reasoning. Our goal is not to claim that ASM is the optimal memory mechanism, but rather to demonstrate that structured intermediate-state representations constitute a promising direction for addressing the memory bottleneck identified by the benchmark.

\section{Experiment Details}
\label{app:experiment}
\subsection{Prompts for Summary and \methodname{} Generation}
\label{app:history_prompt}
Structured prompts are used for both high-level history summarization and the generation of \methodname{} to support effective reasoning over long-duration tasks. These prompts define the action space, normalized coordinate system, and the core decision principles used during interaction. While the history summary prompt focuses on maintaining a compact abstraction of the current task state, the \methodname{} prompt is designed to identify and record semantically significant transitions—such as subgoal achievements or task dependencies—that influence long-term planning. 
% By distilling raw action sequences into these semantic representations, the agent can maintain context without exceeding token limits or losing track of the primary objective. 
The specific structures, field requirements, and generation guidelines for these prompts are detailed in Table ~\ref{table:history_prompt} and Table ~\ref{table:milestone_prompt}.
% \noindent
% Both prompts share the same action schema and a normalized coordinate system (0--1000) to ensure consistent execution across devices.
% The history prompt maintains a compact \texttt{summary\_en} that tracks goal-relevant state, while the milestone prompt additionally outputs \texttt{content\_en} and \texttt{description\_en} to record durable task events that affect future decisions.
% We require exactly one action per step, and update the corresponding fields only when the task state changes materially.
\subsection{Evaluation Protocol and Success Criteria}
\label{app:exp_evaluation_protocol}
This section clarifies how task success is determined in \BenchName{}-Bench and how the evaluation protocol operationalizes long-horizon task completion. The goal is to ensure reproducibility and consistent evaluation across different agent implementations. Rather than relying solely on final UI states, the benchmark evaluates whether agents correctly establish and reuse dependency-critical intermediate states throughout the interaction trajectory.

\paragraph{Task Success Definition.}
Each task in AndroTMem-Bench is annotated with a sparse sequence of state anchors representing task-relevant intermediate states and a final anchor indicating successful completion. A task is considered successful if the agent reaches the final anchor while preserving the required causal dependencies among preceding anchors. This definition reflects the nature of long-horizon GUI workflows, where later decisions often depend on intermediate results obtained several steps earlier.

\paragraph{Dependency Satisfaction.}
The causal dependencies between anchors specify how earlier task states constrain later actions. During evaluation, a dependency is considered satisfied when the downstream behavior of the agent is consistent with the intermediate state required by the task. In practice, this is determined using the semantic content of the annotated anchors together with the observed action trajectory and UI state. For example, if a later step requires reusing a previously extracted value, copied text, selected item, or identified contact, the evaluation verifies that the agent's subsequent actions correctly reflect that earlier state. The evaluation relies on the predefined anchor annotations of each task rather than post-hoc free-form judgment.

\paragraph{Matching and Tolerance.}
Anchor contents may include categorical states, entity references, or structured values. Evaluation follows task-dependent matching rules that preserve the correctness of task-critical information. Exact semantic matching is required for discrete states such as page transitions, subgoal completion, or item selection. For anchors that contain key fields (e.g., prices, contact names, or message recipients), correctness is determined based on whether the extracted value or entity matches the annotated ground-truth state. 
Limited tolerance is applied only to superficial surface variations that do not alter the underlying semantic value, such as formatting differences in numerical values or equivalent textual forms referring to the same entity. Any mismatch in task-critical fields is treated as an incorrect state. This design ensures that evaluation focuses on whether the correct intermediate state is preserved, rather than on character-level formatting differences.

\paragraph{Handling UI Drift and Interruptions.}
The evaluation in \BenchName{}-Bench is performed on pre-recorded interaction trajectories with fixed screenshots and annotated actions. These trajectories may contain interruption-related steps, such as pop-up advertisements, permission dialogs, temporary loading screens, or other unexpected UI states that arise in realistic mobile interaction.
Such cases are not treated as ignorable noise in evaluation. Instead, when an interruption introduces a task-relevant state transition or requires explicit recovery, it is evaluated as part of the trajectory semantics and may be annotated as an \act{exception} anchor or as a dependency-critical intermediate state. In these cases, the agent is required to correctly recognize and handle the interruption; otherwise, the corresponding anchor is considered unsatisfied.
At the same time, the evaluation does not require exact step-by-step trajectory replication. What matters is whether the required semantic state is correctly established and whether the annotated dependency structure is preserved. Therefore, interruption handling is evaluated based on state correctness rather than low-level action identity, but it is not exempted from the success criteria.

\paragraph{Task-Level Metrics.}
Within this evaluation framework, Task Completion Rate (TCR) serves as the primary task-level metric. TCR measures the proportion of tasks in which the agent successfully reaches the final anchor while satisfying the dependency constraints induced by earlier anchors. By incorporating intermediate state consistency into the success definition, TCR captures whether an agent can correctly preserve and reuse task-critical information across long interaction horizons rather than merely reaching the final interface state.

\subsection{Agent Implementation and Model Behavior}
\label{app:exp_model_behavior}
Table~\ref{tab:5_model_performance} and Table~\ref{tab:5_history_ablation_vertical} report relatively low absolute performance for several general-purpose multimodal models, including GPT-4o and GPT-5. Since these results may appear counterintuitive given the strong capabilities of such models in other benchmarks, we clarify the evaluation setup and discuss the main factors affecting performance.

\paragraph{Unified Agent Framework.}
All evaluated models are deployed within the same GUI-agent execution framework. At each step, the model receives the current UI observation (screenshot and UI metadata when available), the user instruction, and the chosen history representation (raw trajectory, summary, or ASM). The model is then required to produce a single structured action following the predefined action schema of AndroTMem-Bench. This design ensures that differences in performance primarily reflect the model's ability to interpret the UI state and maintain task-relevant information across steps rather than differences in system implementation.

\paragraph{Action-Schema Alignment.}
The action space in AndroTMem-Bench is defined using a normalized coordinate system and a fixed set of action types. To ensure compatibility across models, the output format is strictly constrained using structured JSON schemas together with automatic retry mechanisms when formatting errors occur. As a result, the evaluation pipeline minimizes failures caused by output formatting mismatches or tool-interface inconsistencies.

\paragraph{Long-Horizon Interaction Difficulty.}
Despite their strong reasoning abilities, general-purpose multimodal models are not specifically optimized for long-horizon GUI interaction. Tasks in AndroTMem-Bench often involve 30–60 interaction steps with cross-app dependencies and intermediate state reuse. In our experiments, many failures occur when models must recall or reuse information extracted several steps earlier (e.g., previously observed values, selected entities, or copied content). These errors accumulate over long trajectories and significantly reduce overall task completion rates.

\paragraph{Failure Patterns Observed in Large Models.}
Qualitative inspection of trajectories reveals several common failure modes for large general-purpose models, including losing track of previously extracted values, misbinding entities across applications, or drifting from the intended subgoal after multiple steps. These behaviors are consistent with the memory-related failure modes discussed in Section~\ref{sec:6_failure_modes}. Importantly, these failures occur even when individual perception or reasoning steps appear correct, suggesting that the primary difficulty arises from maintaining consistent intermediate state over long interaction horizons.

\paragraph{Interpretation of Absolute Performance.}
Therefore, the relatively low absolute performance of certain large models should not be interpreted as a limitation of the models themselves, but rather as an indication of the difficulty of long-horizon GUI interaction under strict action-execution constraints. In this sense, AndroTMem-Bench exposes a gap between strong multimodal reasoning capabilities and reliable long-horizon interaction memory, which is precisely the phenomenon the benchmark is designed to diagnose and measure.

\clearpage
~% % --- 颜色定义 ---
\definecolor{boxblue}{RGB}{7, 71, 126} 
\definecolor{boxlight}{HTML}{EBF5FF}

\tcbset{
    summarystyle/.style={
        enhanced,
        breakable, % 允许跨页的关键
        colback=boxlight,
        colframe=boxblue,
        coltitle=white,
        fonttitle=\bfseries\sffamily,
        arc=5pt,
        boxrule=1pt,
        left=10pt, right=10pt, top=8pt, bottom=8pt,
        title={prompt},
        before upper={\parindent0pt}, 
    }
}

% --- 正文部分 ---

% 使用 captionof 定义标题，这样它就不会被当作浮动体处理，会紧跟文字出现
\vspace{1em}

% \begin{minipage}{\linewidth}
{\captionsetup{hypcap=false}%
\captionof{table}{Detailed Prompt Structures for History}\label{table:history_prompt}}%

\vspace{-2mm}

\begin{tcolorbox}[summarystyle, title=Prompt Details,title after break={Prompt Details (continued)}]

    \small
    Mobile Assistant.
    
    \vspace{0.8em}
    {\large\textbf{I. Action Space}} \vspace{0.4em} \\
    {\raggedright Allowed actions: {\footnotesize\ttfamily \textbf{[}%
    \{"label":"Tap","value":"tap"\},\allowbreak
    \{"label":"Input Text","value":"text"\},\allowbreak
    \{"label":"Need Feedback","value":"need\_feedback"\},\allowbreak
    \{"label":"Long Press","value":"long\_press"\},\allowbreak
    \{"label":"Swipe","value":"swipe"\},\allowbreak
    \{"label":"Swipe (Two Points)","value":"swipe\_two\_points"\},\allowbreak
    \{"label":"Wait","value":"wait"\},\allowbreak
    \{"label":"Finish","value":"FINISH"\},\allowbreak
    \{"label":"Open App","value":"open\_app"\},\allowbreak
    \{"label":"Capture Screen","value":"capture\_screen"\},\allowbreak
    \{"label":"Home","value":"home"\},\allowbreak
    \{"label":"Back","value":"back"\}\textbf{]}}\par}

    \vspace{0.8em}
    {\large\textbf{II. Field Requirements}} \vspace{0.4em} \\
    1. action.action: String. Must be in Action Space. \\
    2. action.x,y,x\_end,y\_end: Use normalized coords 0-1000. (0,0)=Top-Left. (1000,1000)=Bottom-Right. For "tap"/"long\_press" use (x,y). For "swipe\_two\_points" use start(x,y) \& end(x\_end,y\_end). Unused=0. \\
    3. action.value: String. Input text, feedback request, or app name. \\
    4. action.direction: String. "up"|"down"|"left"|"right" (for "swipe"). \\
    5. action.distance: String. "long"|"medium"|"short" (for "swipe").

    \vspace{0.8em}
    {\large\textbf{III. Decision Principles}} \vspace{0.4em} \\
    Visual Evidence: Only act on what you see. If loading, use "wait". \\
    Precision: Target UI element center. Use 0-1000 scale carefully. \\
    Step-by-Step: Output ONE action per response.

    \vspace{0.8em}
    {\large\textbf{IV. Summary Generation Guidelines}} \vspace{0.4em} \\
    The summary\_en field represents a compact, high-level abstraction of the current task progress and interface state. It serves as persistent context for subsequent decision steps.

    \vspace{0.4em}
    When writing summary\_en: \\
    1. Describe the current goal-relevant state of the task, not low-level UI details. \\
    2. Capture what has been accomplished so far and what remains unresolved. \\
    3. Include critical constraints, user inputs, or system feedback that affect future actions. \\
    4. Avoid step-by-step narration or coordinate-level descriptions. \\
    5. Keep the summary concise, factual, and stable across steps unless the task state meaningfully changes.

    \vspace{0.4em}
    The summary should enable future steps to reason about progress without access to full action history.

    \vspace{0.8em}
    {\large\textbf{V. Output Format}}\par\vspace{0.4em}
    Return a single JSON object:\par
    % wujie revised
    % {\footnotesize\ttfamily \{ "action": \{ "action": "...", "x": 0, "y": 0, "value": "", "x\_end": 0, "y\_end": 0 \}, "summary\_en": "..." \}}
    {\sloppy
    {\footnotesize\ttfamily
    \{ "action": \{ "action": "...", "x": 0, "y": 0, "value": "", "x\_end": 0, "y\_end": 0 \}, "summary\_en": "..." \}}
    \par}

\end{tcolorbox}
% \end{minipage}

\par\vspace{0.5em}
~% % --- 颜色定义 ---
\definecolor{boxgreen}{RGB}{34, 139, 34}    % 深森林绿
\definecolor{greenlight}{RGB}{245, 252, 245} % 极浅绿色

\tcbset{
    milestone/.style={
        enhanced,
        breakable,
        colback=greenlight,
        colframe=boxgreen,
        coltitle=white,
        fonttitle=\bfseries\sffamily,
        arc=5pt,
        boxrule=1pt,
        left=10pt, right=10pt, top=8pt, bottom=8pt,
        title={prompt},
        before upper={\parindent0pt},
    }
}

    % \caption{Detailed Prompt Structures for Milestones}
    % \vspace{-4mm}
    % \begin{tcolorbox}[milestone, title=Prompt Details]
    \vspace{1em}
    
    % \begin{minipage}{\linewidth}
    {\captionsetup{hypcap=false}%
    \captionof{table}{Detailed Prompt Structures for \methodname{}}\label{table:milestone_prompt}}%
    \vspace{-2mm}
    
    \begin{tcolorbox}[milestone, title=Prompt Details,title after break={Prompt Details (continued)}] % 不要 breakable
    \small
    Mobile Assistant.

    \vspace{0.8em}
    {\large\textbf{I. Action Space}} \vspace{0.4em} \\
    % 使用 \raggedright 防止两端对齐拉伸，并在关键位置加 \allowbreak
    {\raggedright Allowed actions: {\footnotesize\ttfamily \textbf{[}\{"label":"Tap","value":"tap"\},\allowbreak \{"label":"Input Text","value":"text"\},\allowbreak \{"label":"Need Feedback","value":"need\_feedback"\},\allowbreak \{"label":"LongPress","value":"long\_press"\},\allowbreak \{"label":"Swipe","value":"swipe"\},\allowbreak \{"label":"Swipe (Two Points)","value":"swipe\_two\_points"\},\allowbreak \{"label":"Wait","value":"wait"\},\allowbreak \{"label":"Finish","value":"FINISH"\},\allowbreak \{"label":"Open App","value":"open\_app"\},\allowbreak \{"label":"Capture Screen","value":"capture\_screen"\},\allowbreak \{"label":"Home","value":"home"\},\allowbreak \{"label":"Back","value":"back"\}\textbf{]}} \par}

    \vspace{0.8em}
    {\large\textbf{II. Field Requirements}} \vspace{0.4em} \\
    1. action.action: String. Must be in Action Space. \\
    2. action.x,y,x\_end,y\_end: Use normalized coords 0-1000. (0,0)=Top-Left. (1000,1000)=Bottom-Right. For "tap"/"long\_press" use (x,y). For "swipe\_two\_points" use start(x,y) \& end(x\_end,y\_end). Unused=0. \\
    3. action.value: String. Input text, feedback request, or app name. \\
    4. action.direction: String. "up"|"down"|"left"|"right" (for "swipe"). \\
    5. action.distance: String. "long"|"medium"|"short" (for "swipe").

    \vspace{0.8em}
    {\large\textbf{III. Decision Principles}} \vspace{0.4em} \\
    Visual Evidence: Only act on what you see. If loading, use "wait". \\
    Precision: Target UI element center. Use 0-1000 scale carefully. \\
    Step-by-Step: Output ONE action per response.

    \vspace{0.8em}
    {\large\textbf{IV. Anchor and Causal Structure Generation Guidelines}} \vspace{0.4em} \\
    State anchors represent semantically meaningful task events that influence long-term planning and decision-making. 
    Each anchor summarizes a key transition, dependency, or achievement. 
    Beyond recording important states, the model should explicitly identify causal links between anchors so that the interaction history is organized as a structured dependency graph rather than a flat list of events.
    
    \vspace{0.4em}
    Anchor categories: \\
    \textbf{[}subgoal\textbf{]} Achievement of an intermediate objective. \\
    \textbf{[}state\_change\textbf{]} Entry into a new screen, mode, or functional state. \\
    \textbf{[}dependency\textbf{]} Completion of a prerequisite required for future steps. \\
    \textbf{[}exception\textbf{]} Errors, failures, or unexpected states requiring handling. \\
    \textbf{[}context\_info\textbf{]} Important parameters, settings, or user-provided information. \\
    \textbf{[}finish\textbf{]} Final goal successfully completed.
    
    \vspace{0.4em}
    Causal link types: \\
    \textbf{[}prerequisite\textbf{]} A previous anchor must hold before the current one can occur. \\
    \textbf{[}enables\textbf{]} A previous anchor creates the condition for a future action or subgoal. \\
    \textbf{[}result\_of\textbf{]} The current anchor is the direct result of a previous anchor or action outcome. \\
    \textbf{[}blocks\textbf{]} An exception or state prevents progress until resolved.
    
    \vspace{0.4em}
    When generating anchors and links: \\
    1. content\_en should be a concise, category-tagged semantic statement describing the current anchor. \\
    2. description\_en should explain why this anchor matters for subsequent reasoning and execution. \\
    3. If the current anchor has a direct causal dependency on a previous anchor, generate a causal link identifying the source anchor and relation type. \\
    4. Only create causal links for decision-critical dependencies, not for trivial temporal succession. \\
    5. Only generate anchors for durable and task-relevant events; avoid trivial UI transitions. \\
    6. Do not repeat previous anchors unless the task state fundamentally changes. \\
    7. Preserve both key intermediate states and the causal structure connecting them.

    \vspace{0.8em}
    {\large\textbf{V. Output Format}} \vspace{0.4em} \\
    Return a single JSON object: \\
    {\footnotesize\ttfamily \{ "action": \{ "action": "...", "x": 0, "y": 0, "value": "", "x\_end": 0, "y\_end": 0 \}, "content\_en": "...", "description\_en": "..." ,"causal\_link": \{ "source": "...", "relation": "..." \}\}}
    \end{tcolorbox}
    % \end{minipage}
    % \label{table:milestone_prompt}
\FloatBarrier
~\begin{figure*}[ht]
    \centering
    \includegraphics[width=\textwidth]{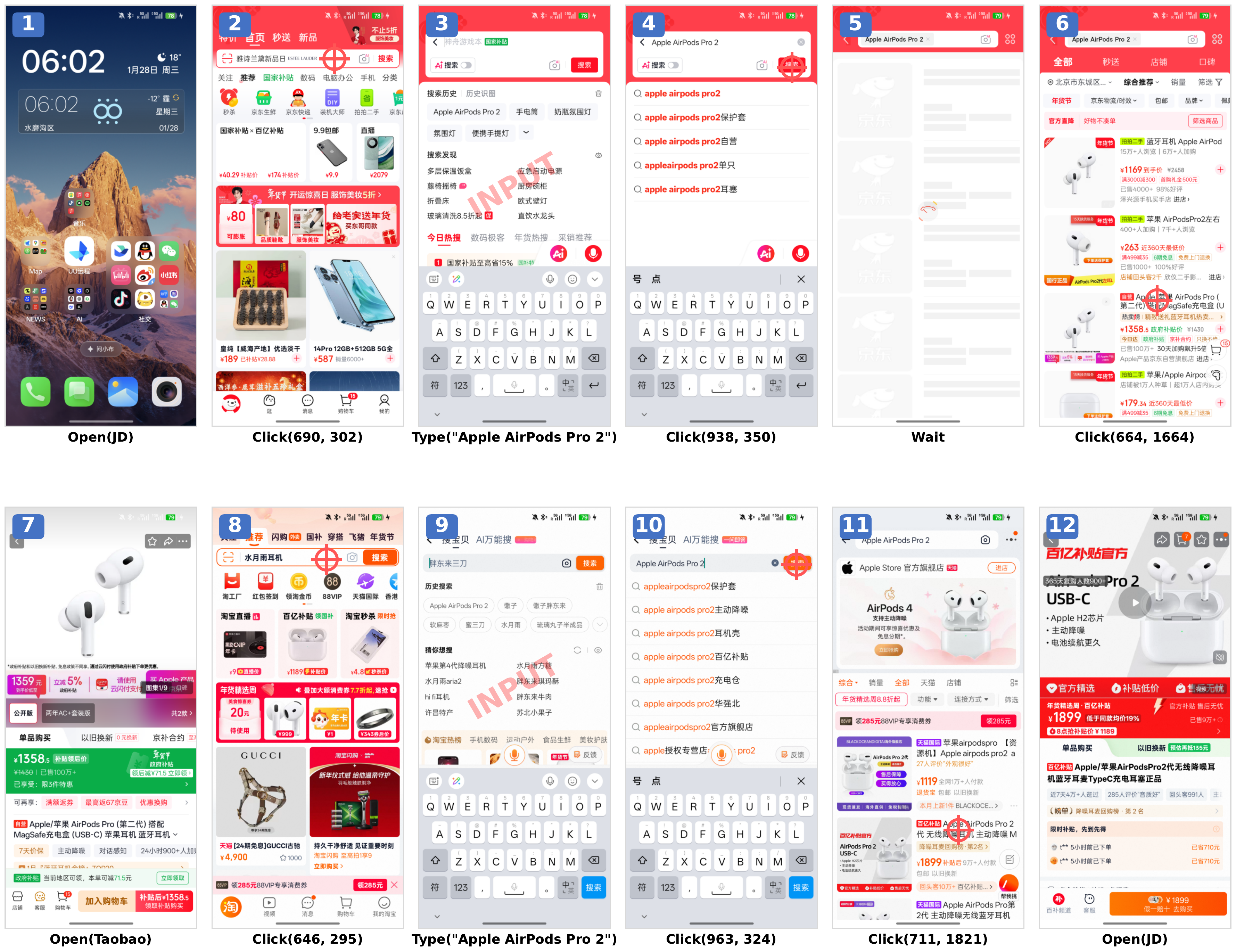}
    \caption{\textbf{A complete trajectory example from \BenchName{}.}}
    \label{fig:appendix_trajectory_part1}
\end{figure*}
~\begin{figure*}[ht]
    \centering
    \includegraphics[width=\textwidth]{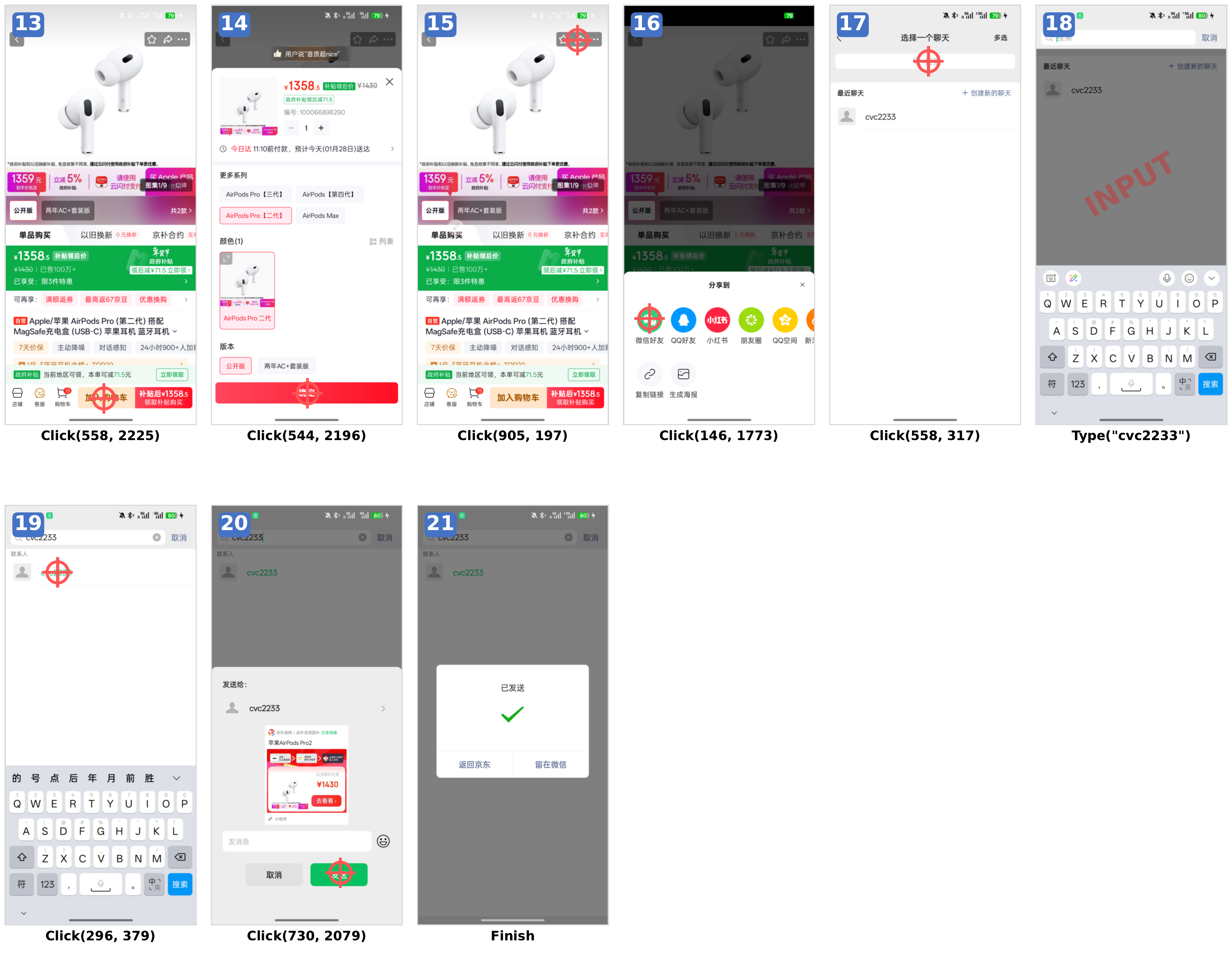}
    \caption{\textbf{A complete trajectory example from \BenchName{}.}}
    \label{fig:appendix_trajectory_part2}
\end{figure*}
\end{document}